\documentclass[journal]{IEEEtran}

\usepackage{cite}
\usepackage{graphicx}
\usepackage{amsmath,amssymb}
\usepackage{booktabs}
\usepackage{multirow}
\usepackage{array}
\usepackage{tabularx}
\usepackage{url}
\usepackage{xcolor}
\usepackage[caption=false,font=footnotesize]{subfig}
\usepackage{pifont}

\usepackage{url}
\usepackage{xcolor}
\usepackage[caption=false,font=footnotesize]{subfig}
\usepackage{pifont}

\usepackage[hidelinks]{hyperref}
\newcommand{\cmark}{\ding{51}}
\newcommand{\xmark}{\ding{55}}

\begin{document}

\title{SGWIB: Sliced Gromov--Wasserstein Information Bottleneck
for Video Highlight Detection}

\author{
Hanjuan~Huang,
Yung-Chieh~Yeh,
and Hsing-Kuo~Pao%
\thanks{Hanjuan Huang is with the Key Laboratory of Agricultural
Machinery Intelligent Control and Manufacturing of Fujian Province
University, College of Mechanical and Electrical Engineering,
Wuyi University, Wuyishan 354300, China
(e-mail: huanghanjuan@wuyiu.edu.cn).}%
\thanks{Yung-Chieh Yeh and Hsing-Kuo Pao are with the National Taiwan
University of Science and Technology, No. 43, Sec. 4, Keelung Rd.,
Taipei, Taiwan
(e-mail: M11115084@mail.ntust.edu.tw; pao@mail.ntust.edu.tw).}%
\thanks{Corresponding author: Hsing-Kuo Pao
(e-mail: pao@mail.ntust.edu.tw).}
}

\maketitle

\begin{abstract}
Video highlight detection aims to identify temporally important
segments that capture the most informative or engaging events in a
video. Reliable prediction therefore requires not only discriminative
segment representations but also preservation of the temporal
relationships among neighboring and distant segments. The information
bottleneck principle has proven effective for learning compact and
task-relevant representations, yet it has not been explored for video
highlight detection, and applying conventional formulations directly
would overlook inter-segment relational structure and distort
highlight-relevant temporal organization during compression. We
therefore introduce the Sliced Gromov--Monge Gap (SGMG), a
structure-aware regularizer that measures the excess relational
distortion induced by a prescribed source-to-bottleneck mapping
relative to an optimal sliced structural correspondence. Building on
SGMG, we develop SGWIB, an information-bottleneck framework for
single-modal video highlight detection that learns compact bottleneck
representations while preserving inter-segment temporal structure. We
further introduce Home--Away-Related Contextual Pseudo-Labels and a
contextual disentanglement module that reduce sports-specific
contextual bias by separating highlight-oriented information from
contextual patterns. Experiments on MrHiSum and MoSu show that SGWIB
attains the best Kendall's $\tau$, Spearman's $\rho$,
$\mathrm{mAP}@50$, and $\mathrm{mAP}@30$ among the compared
single-modal methods on both datasets. On MrHiSum, the visual model
improves the strongest previous results by $0.031$, $0.031$, $0.87$,
and $0.75$ on these four metrics, respectively. These results show
that structure-aware information-bottleneck regularization combined
with contextual disentanglement improves segment-level highlight
prediction.
\end{abstract}


\section{Introduction}
\label{sec:introduction}

Video highlight detection (VHD) aims to identify temporally important
segments and assign each segment an importance score, supporting
applications such as video browsing, summarization, retrieval, and
content recommendation. The importance of a segment depends not only
on its local content but also on its relation to surrounding segments.
Effective VHD therefore requires both discriminative segment
representations and reliable preservation of temporal relational
structure.

Recent studies have advanced VHD through online prediction, test-time
adaptation, hierarchical temporal modeling, multimodal interaction, and
pretrained vision--language models~\cite{chang2025aha,
islam2025highlighttta,beedu2026hiersum,um2025keyword,
qiu2026tvhighlights,momentmamba2026}. Two challenges nevertheless
remain insufficiently explored. First, learned representations may
retain redundant or task-irrelevant contextual information. Sports
videos make this especially visible: crowd responses, venue acoustics,
camera activity, and broadcast patterns correlate with highlight events
without providing stable evidence of event importance. Second,
compressing high-dimensional video representations without explicitly
accounting for inter-segment relations can distort the temporal
organization needed to separate salient events from their surrounding
context.

The information bottleneck (IB) principle offers a natural mechanism
for learning compact representations that retain task-relevant
information while suppressing unnecessary dependence on the
input~\cite{tishby2000information}. IB has been explored in related
video and sequential representation-learning
tasks~\cite{srivastava2021vib,fan2022dtr,zhong2023semantic,
li2023robust}, but to our knowledge it has not been systematically
investigated for VHD. More importantly, conventional variational IB
methods approximate the compression term by the Kullback--Leibler
divergence between the encoder posterior and a predefined
prior~\cite{alemi2017deep}. Such prior matching is computationally
cheap, yet it does not explicitly preserve the relational geometry
among temporal segments, and under strong compression it may alter both
local neighborhoods and global temporal organization. The
Gromov--Wasserstein information bottleneck (GWIB) offers a
structure-aware alternative by relating empirical kernelized dependence
to the Gromovized Monge gap~\cite{yang2024gwib}, but full
Gromov--Wasserstein optimization is prohibitively expensive for long
and high-dimensional video sequences.

To resolve this structure--efficiency trade-off, we introduce the
Sliced Gromov--Monge Gap (SGMG), a sliced approximation of the
Gromovized Monge gap that measures the excess relational distortion
induced by a prescribed source-to-bottleneck mapping relative to an
optimal sliced structural correspondence. Building on SGMG, we develop
SGWIB, a Sliced Gromov--Wasserstein Information Bottleneck framework
for single-modal VHD. SGWIB jointly learns a stochastic bottleneck
representation and a segment-level highlight predictor, encouraging
compact task-relevant representations while preserving inter-segment
temporal organization. By exploiting the closed-form one-dimensional
sliced Gromov--Wasserstein optimum over multiple fixed random
projections, together with normalized temporal coordinates and validity
masking, SGMG remains tractable where full GW optimization is not. We
further establish an information-theoretic interpretation in which the
expected SGMG controls an upper bound on empirical sliced kernelized
dependence, without claiming an exact bound on Shannon mutual
information.

To reduce sports-specific contextual bias, we further introduce
Home--Away-Related Contextual Pseudo-Labels (HAR-CPLs) and an
HAR-CPL-guided Contextual Disentanglement Module (HAR-CDM). HAR-CPLs
provide weak video-level contextual supervision for sports videos,
while HAR-CDM separates the backbone representation into task-oriented
and context-oriented components. Orthogonality regularization applies
to all valid videos to reduce redundancy between the two branches,
whereas HAR-CPL-based pseudo-environment supervision is activated only
for sports videos. SGWIB then regularizes the structural compression of
the task-oriented representation, again for all valid videos. The two
modules therefore act at different levels: HAR-CDM reduces unstable
contextual correlations, while SGMG preserves task-relevant temporal
structure during bottleneck compression. The visual and audio models
follow the same formulation but are trained independently, without
cross-modal interaction.

Experiments on MrHiSum and MoSu show that SGWIB consistently improves
temporal importance ranking and highlight retrieval. On the primary
visual branch, it attains the best Kendall's $\tau$, Spearman's $\rho$,
$\mathrm{mAP}@50$, and $\mathrm{mAP}@30$ among the compared methods on
both datasets. Ablation studies confirm the complementary effects of
HAR-CDM and SGWIB, while comparisons with KL-based bottleneck
regularization show more favorable task-level performance and
substantially stronger structural preservation at modest additional
training cost.

The main contributions are summarized as follows:

\begin{itemize}

\item
We introduce SGMG, a mapping-aware structural regularizer that
quantifies the excess relational distortion induced by a prescribed
source-to-bottleneck mapping relative to an optimal sliced structural
correspondence.

\item
We develop SGWIB, an information-bottleneck framework for single-modal
VHD. By combining stochastic bottleneck learning with time-aware and
validity-mask-aware SGMG regularization, it provides a structure-aware
alternative to KL-based prior matching while avoiding the cost of full
GW optimization.

\item
We provide an information-theoretic interpretation of SGMG by showing
that its expectation controls an upper bound on empirical sliced
kernelized dependence, supporting its use as a tractable structural
compression regularizer.

\item
We introduce HAR-CPL-guided contextual disentanglement to reduce
sports-specific contextual bias, and show through extensive experiments
that HAR-CDM and SGWIB bring complementary benefits in highlight
prediction, structural preservation, and computational efficiency.

\end{itemize}
\section{Related Work}
\label{sec:related_work}

Video highlight detection aims to identify temporally salient segments
that capture important events within a video. Early and representative
video summarization methods mainly focused on modeling temporal
importance from visual features. VASNet employs self-attention to
capture long-range temporal dependencies~\cite{fajtl2019summarizing},
while PGL-SUM combines global and local attention for more effective
temporal modeling~\cite{apostolidis2021combining}. Generative
approaches such as SUM-GAN~\cite{mahasseni2017unsupervised} and
AC-SUM-GAN~\cite{apostolidis2021ac} learn compact summaries through
adversarial reconstruction, whereas iPTNet jointly models temporal
dependencies for importance prediction~\cite{jiang2022joint}.

More recent studies have improved VHD from the perspectives of
cross-category generalization, feature decomposition, contextual
modeling, and pretrained representation transfer. The SL-module
promotes cross-category highlight learning through representation
decomposition and semantic alignment~\cite{xu2021crosscategory},
while related feature decomposition strategies further separate
highlight-relevant information from category-specific variations
~\cite{zhang2023crosscategory}. CSTA strengthens contextual temporal
modeling for highlight prediction~\cite{son2024csta}, and SummDiff
introduces diffusion-based representation learning for video
summarization~\cite{kim2025summdiff}. Highlight-CLIP further exploits
transferable knowledge from pretrained vision--language models
~\cite{han2024highlightclip}, while unsupervised audio--visual
recurrence has been used to construct pseudo-highlight supervision
without manual annotations~\cite{islam2025recurrence}.

Multimodal and query-conditioned approaches additionally exploit
visual, audio, and textual information through video--text matching,
contrastive learning, and cross-modal interaction
~\cite{xiong2023dualstream,jiang2024mctvhd,
sun2024trdetr,xiao2024uvcom}. More recent methods employ Video-LLMs,
context-aware alignment, LLM-generated supervision, or test-time
adaptation to improve semantic understanding and generalization
~\cite{yang2025timeexpert,islam2026highlighttta,
qiu2026tvhighlights,moon2026cva}.

Despite these advances, existing VHD methods primarily focus on
enhancing feature representations, temporal modeling, semantic
alignment, generative summarization, or multimodal interaction,
whereas the effect of representation compression on inter-segment
temporal relational structure remains largely unexplored. Our work
addresses this complementary problem through a structure-aware
information bottleneck tailored to VHD. Specifically, HAR-CDM
separates task-oriented representations from sports-related contextual
variations, while SGWIB regularizes the source-to-bottleneck
transformation to preserve inter-segment relational structure during
compression.

\subsection{Information Bottleneck and Structural Dependence}

The information bottleneck (IB) principle seeks a compact
representation that preserves target-relevant information while
discarding redundant dependence on the input
~\cite{tishby2000information}. Its variational formulation enables
end-to-end stochastic optimization through reparameterization
~\cite{alemi2017deep}, and has been applied to video action
recognition, temporal representation learning, multimodal learning,
and other representation-learning tasks
~\cite{srivastava2021vib,liu2024timex,pan2024rmib,
almudevar2025aligning,zhang2025vibdet}. However, conventional
variational IB typically approximates the compression term through the
KL divergence between the encoder posterior and a predefined prior,
which constrains latent capacity but does not explicitly preserve the
relational geometry induced by the bottleneck mapping.

Alternative dependence measures have therefore been developed to avoid
or complement KL-based compression. MINE estimates mutual information
through a neural variational lower bound~\cite{belghazi2018mine}, CLUB
constructs a contrastive upper bound for mutual-information
minimization~\cite{cheng2020club}, and the Conditional Entropy
Bottleneck reformulates compression according to the minimum necessary
information principle~\cite{fischer2020ceb}. Optimal-transport-based
approaches further provide geometry-aware dependence measures, such as
the Wasserstein Dependency Measure~\cite{ozair2019wasserstein}.
More recently, GWIB relates empirical kernelized dependence to the
Gromovized Monge gap, introducing structural information into
bottleneck regularization~\cite{yang2024gwib}. Nevertheless, full
Gromov--Wasserstein optimization remains computationally expensive for
long video sequences.

Different from these methods, we introduce the Sliced Gromov--Monge
Gap (SGMG) as a mapping-aware structural compression surrogate for
single-modal video representations. SGMG measures the excess
relational distortion induced by the prescribed mapping from
$\widetilde{\mathbf X}^{m}$ to $\mathbf Z^{m}$ relative to an optimal
one-dimensional sliced structural correspondence. By exploiting the
identity or reversed correspondence of one-dimensional SGW
~\cite{vayer2019sliced} and aggregating the resulting gap over multiple
random projections, SGMG provides a computationally tractable means of
preserving temporal relational structure. It is used as a
structure-aware compression surrogate rather than as an exact estimator
or universal variational upper bound of Shannon mutual information
$I(\widetilde{\mathbf X}^{m};\mathbf Z^{m})$.
\section{Methodology}
\label{sec:method}

\subsection{Problem Definition and Notation}
\label{subsec:problem_notation}

\subsubsection{Problem Definition}
\label{subsubsec:problem_definition}

Given a video represented as a sequence of temporal segments, video
highlight detection assigns a continuous importance score to each
valid segment. Let
$\mathcal{V}_{\mathrm{train}}=\{1,\ldots,N\}$ denote the training-video
index set. The training set is
\begin{equation}
\mathcal{D}_{\mathrm{train}}
=
\left\{
\left(
\widetilde{\mathbf X}_{b}^{v},
\widetilde{\mathbf X}_{b}^{a},
\mathbf y_b,
\mathbf M_b,
c_b
\right)
\right\}_{b=1}^{N},
\label{eq:training_set}
\end{equation}
where
$\widetilde{\mathbf X}_{b}^{v}$ and
$\widetilde{\mathbf X}_{b}^{a}$ are the padded visual and audio
sequences,
$\mathbf y_b\in\mathbb R^{L}$ is the segment-level highlight-score
sequence,
$\mathbf M_b\in\{0,1\}^{L}$ is the temporal validity mask, and
$c_b\in\{0,1\}$ indicates whether the video belongs to a sports
category.

For modality $m\in\mathcal M=\{v,a\}$, an independently optimized
modality-specific model predicts
$\widehat{\mathbf y}_{b}^{m}\in\mathbb R^{L}$.
The objective is to accurately estimate highlight scores at valid
temporal positions while suppressing task-irrelevant contextual
information and limiting distortion of temporal relational structure.

\subsubsection{Notation}
\label{subsubsec:notation}

The principal notation is summarized in
Table~\ref{tab:main_notation}. The video index $b$ is omitted in
single-video module-level formulations and restored for dataset-level
statistics and mini-batch aggregation.

\begin{table}[t]
\centering
\caption{Summary of the principal notation.}
\label{tab:main_notation}
\footnotesize
\renewcommand{\arraystretch}{1.06}
\setlength{\tabcolsep}{2pt}
\begin{tabularx}{\columnwidth}{
    @{}
    p{0.27\columnwidth}
    @{\hspace{2pt}}
    X
    @{}
}
\toprule
\centering\textbf{Symbol} & \textbf{Description} \\
\midrule

$\displaystyle m\in\mathcal M=\{v,a\}$
&
Modality index for visual ($v$) and audio ($a$).
\\

$\displaystyle
\widetilde{\mathbf X}^{m}\in\mathbb R^{L\times d_m}$
&
Padded source sequence; also the detached source-side
geometry reference of SGMG.
\\

$\displaystyle \mathbf M,\;\mathcal I,\;n$
&
Temporal validity mask, valid index set, and $n=|\mathcal I|$.
\\

$\displaystyle \mathbf y,\widehat{\mathbf y}^{m}$
&
Ground-truth and predicted highlight-score sequences.
\\

$\displaystyle \mathbf X^{m}$
&
Backbone representation.
\\

$\displaystyle
\mathbf H^{m},
\mathbf X_{\mathrm{spu}}^{m}$
&
Task-oriented and context-oriented representations
from HAR-CDM.
\\

$\displaystyle \mathbf s^{m},\;\mathbf Z^{m}$
&
Stochastic channel scale shared over time, and the
resulting bottleneck representation.
\\

$\displaystyle c,\;s$
&
Sports-category indicator and sports-supervision mask.
\\

$\displaystyle e^{m},\;\tau^{m},\;y^{\mathrm{env},m}$
&
Contextual activity measure, its median threshold, and the
binary pseudo-environment label.
\\

$\displaystyle
\boldsymbol{\theta}_{p}^{m}\in\mathbb R^{d_m+1}$
&
Fixed unit projection direction, $p=1,\ldots,P$.
\\

$\displaystyle
C_{\mathrm{map}}^{m,(p)},
C_{\mathrm{SGW}}^{m,(p)}$
&
Mapping cost and sliced structural reference of slice $p$.
\\

$\displaystyle \beta_m,\;\gamma_m$
&
Weights of $\mathcal L_{\mathrm{SGMG}}^{m}$ and of the two HAR-CDM
terms in Eq.~\eqref{eq:total_training_loss}.

\\

\bottomrule
\end{tabularx}
\end{table}

\subsection{Framework Overview}
\label{subsec:overview}

\begin{figure*}[t]
  \centering
  \includegraphics[width=0.83\textwidth]{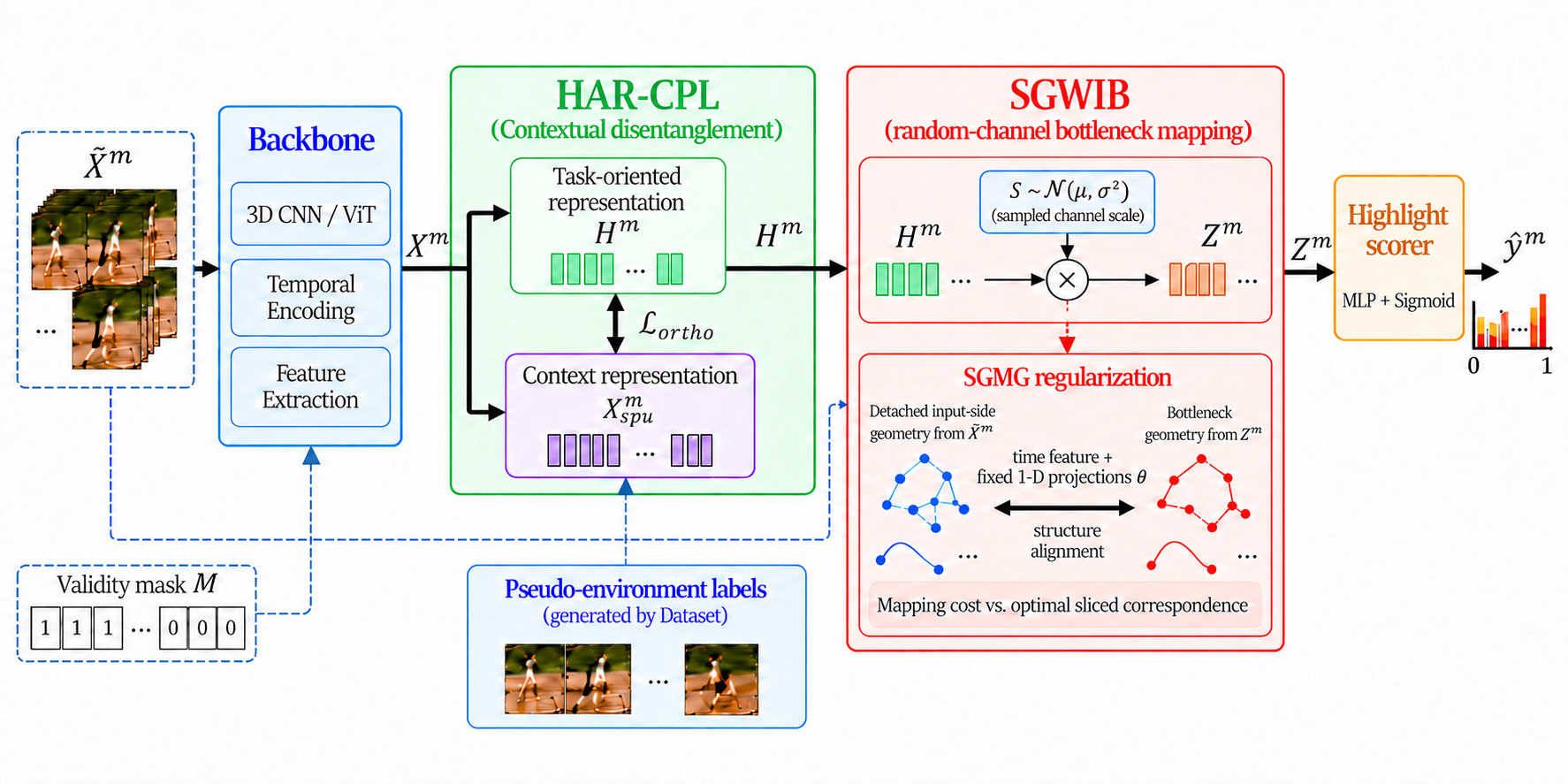}
  \caption{Overview of the proposed SGWIB framework. The visual and
  audio branches follow the same formulation but are optimized
  independently.}
  \label{fig:sgwib_framework}
\end{figure*}

Figure~\ref{fig:sgwib_framework} illustrates SGWIB, consisting of a
modality-specific backbone, HAR-CDM, an SGMG-regularized stochastic
bottleneck, and a segment-level highlight predictor. The backbone
extracts temporally aligned features from the padded sequence.
HAR-CDM then decomposes them into task-oriented and context-oriented
representations using orthogonality regularization and
sports-specific pseudo-environment supervision. The task-oriented
representation is passed through a multiplicative stochastic channel
that preserves the feature dimension, while SGMG compares the detached
source geometry $\widetilde{\mathbf X}^{m}$ with the bottleneck
geometry over multiple one-dimensional projections to constrain
compression-induced relational distortion. Finally, the bottleneck
representation is mapped to segment-level highlight scores. Visual and
audio branches follow the same formulation but use separate parameters
and are optimized independently.

\subsection{Backbone Module}
\label{subsec:backbone}

For modality $m\in\mathcal M$, the backbone receives the padded source
sequence $\widetilde{\mathbf X}^{m}$ and validity mask $\mathbf M$:
\begin{equation}
\mathbf X^{m}
=
E_m
\left(
\widetilde{\mathbf X}^{m},
\mathbf M
\right),
\qquad
\mathbf X^{m}\in\mathbb R^{L\times d_m},
\label{eq:backbone_encoding}
\end{equation}
where $L$ is the padded sequence length and $d_m$ is the feature
dimension. The backbone preserves the input dimension, so
$\widetilde{\mathbf X}^{m}$ and all downstream representations share
the same feature space, which allows SGMG to compare the input and
bottleneck geometries directly. The mask entry $M_t=1$ denotes a valid
temporal segment and $M_t=0$ denotes padding. The same validity mask is
used in temporal feature extraction, geometry construction, and loss
computation so that padded positions do not affect optimization.

The temporal mask $\mathbf M$ is distinct from the sports-supervision
mask $s$ introduced below: $\mathbf M$ handles variable-length
sequences, whereas $s$ determines whether a video contributes to
pseudo-environment supervision.
\subsection{HAR-CPL-Guided Contextual Disentanglement}
\label{subsec:har_cpl_har_cdm}
Given $\mathbf X^{m}$, HAR-CDM separates information useful for
highlight prediction from contextual variations that may be
spuriously correlated with highlights. This issue is particularly
relevant to sports videos, where crowd responses, venue acoustics,
camera activity, and broadcast behavior may correlate with salient
events without providing stable evidence of event importance.
We therefore construct modality-specific
Home--Away-Related Contextual Pseudo-Labels (HAR-CPLs) as weak
video-level supervision for the context-oriented branch.

Our disentanglement follows the causal attention formulation of
CAAM~\cite{wang2021causal}, which partitions a self-attention map into
causal and confounding streams. Whereas CAAM discovers the partition
through self-annotation driven by task labels alone, we follow
\cite{yu2025causal} in supplying explicit pseudo-environment labels as
a weak supervisory signal, since task labels by themselves cannot
expose the home--away contextual factor in our setting. Unlike
\cite{yu2025causal}, which removes environment-related information
outright, we retain it and only indicate which component carries a
large contextual load, because in highlight detection such factors are
not always causally irrelevant to salient events.

\subsubsection{Sports-Video Selection and Pseudo-Environment Labels}
Let $c_b=1$ indicate a sports video. The sports subset and supervision
mask are
\begin{equation}
\begin{aligned}
\mathcal S_{\mathrm{sports}}
&=
\left\{
b\in\mathcal V_{\mathrm{train}}
\mid c_b=1
\right\},
\\
s_b
&=
\mathbb I
\left[
b\in\mathcal S_{\mathrm{sports}}
\right]
=
c_b.
\end{aligned}
\label{eq:sports_subset_and_mask}
\end{equation}
For sports video $b$, let $\mathcal I_b$ denote its valid temporal
indices, $n_b=|\mathcal I_b|$, and
$\{\widetilde{\mathbf x}_{b,t}^{m}\}_{t\in\mathcal I_b}$ its valid
segment features. Contextual activity is measured differently for the
two modalities: audio features carry acoustic intensity directly in
their magnitude, whereas the magnitude of visual features reflects
scene content rather than activity, so temporal variation is the more
informative signal. We therefore use root-mean-square feature energy
for audio and the average variation between adjacent valid features
for video:
\begin{equation}
\begin{aligned}
e_b^{a}
&=
\sqrt{
\frac{1}{n_b d_a}
\sum_{t\in\mathcal I_b}
\sum_{j=1}^{d_a}
\left(
\widetilde{x}_{b,t,j}^{a}
\right)^2
},
\\
e_b^{v}
&=
\begin{cases}
\displaystyle
\frac{1}{n_b-1}
\sum_{r=1}^{n_b-1}
\left\|
\widetilde{\mathbf x}_{b,t_{r+1}}^{v}
-
\widetilde{\mathbf x}_{b,t_r}^{v}
\right\|_2,
& n_b\geq2,
\\[4pt]
0,
& n_b<2,
\end{cases}
\\
\tau^{m}
&=
\operatorname{Median}
\left\{
e_b^{m}
\mid
b\in\mathcal S_{\mathrm{sports}}
\right\},
\\
y_b^{\mathrm{env},m}
&=
\mathbb I
\left[
e_b^{m}\geq\tau^{m}
\right].
\end{aligned}
\label{eq:pseudo_environment_generation}
\end{equation}
Here,
$t_1<\cdots<t_{n_b}$ are the ordered valid temporal indices.
The modality-specific threshold $\tau^m$ is estimated only from the
sports subset of the training data. Both activity measures are
non-negative and right-skewed, so the median is used instead of the
mean to keep the two pseudo-environments balanced in size. The binary
label $y_b^{\mathrm{env},m}$ distinguishes relatively low and high
contextual activity; it is neither a true home--away identity nor a
segment-level highlight annotation.

\subsubsection{Contextual Representation Disentanglement}
HAR-CDM decomposes the backbone representation into task-oriented and
context-oriented components:
\begin{equation}
\left(
\mathbf H^{m},
\mathbf X_{\mathrm{spu}}^{m}
\right)
=
\operatorname{HAR\text{-}CDM}_{m}
\left(
\mathbf X^{m},
\mathbf M
\right).
\label{eq:har_cdm_mapping}
\end{equation}
The two components are produced by a shared multi-head attention layer
whose logits are used with opposite signs:
\begin{equation}
\begin{aligned}
\mathbf A_{\mathrm{causal}}^{m}
&=
\operatorname{softmax}
\left(
\frac{\mathbf Q^{m}(\mathbf K^{m})^{\top}}{\sqrt{d_h}}
+
\mathbf B
\right),
\\
\mathbf A_{\mathrm{spu}}^{m}
&=
\operatorname{softmax}
\left(
-\frac{\mathbf Q^{m}(\mathbf K^{m})^{\top}}{\sqrt{d_h}}
+
\mathbf B
\right),
\end{aligned}
\label{eq:har_cdm_attention}
\end{equation}
where $d_h$ is the per-head dimension and $\mathbf B$ is an additive
padding bias that suppresses invalid keys. The two attention maps
share the same value projection $\mathbf V^{m}$, so the branches differ
only in how they aggregate context. The task-oriented branch is
combined with a residual connection before normalization, so that it
retains the original sequence information required for highlight
prediction, whereas the context-oriented branch keeps only the
attended component and is therefore encouraged to carry contextual
rather than task-relevant content.

The context-oriented branch is supervised by the sports-only
pseudo-environment loss, and an orthogonality regularizer penalizes
the squared cosine similarity between the two representations at every
valid position:
\begin{equation}
\begin{aligned}
\mathcal L_{\mathrm{spu}}^{m}
&=
\frac{
\displaystyle
\sum_{b\in\mathcal B}
s_b\,
\ell_{\mathrm{env}}
\left(
\widehat y_b^{\mathrm{env},m},
y_b^{\mathrm{env},m}
\right)
}{
\displaystyle
\sum_{b\in\mathcal B}s_b+\varepsilon
},
\\
\mathcal L_{\mathrm{ortho}}^{m}
&=
\frac{
\displaystyle
\sum_{b\in\mathcal B}
\sum_{t=1}^{L}
M_{b,t}
\left(
\frac{
\left\langle
\mathbf h_{b,t}^{m},
\mathbf x_{\mathrm{spu},b,t}^{m}
\right\rangle
}{
\left\|
\mathbf h_{b,t}^{m}
\right\|_2
\left\|
\mathbf x_{\mathrm{spu},b,t}^{m}
\right\|_2
}
\right)^{2}
}{
\displaystyle
\sum_{b\in\mathcal B}
\sum_{t=1}^{L}
M_{b,t}
+
\varepsilon
},
\end{aligned}
\label{eq:har_cdm_objective}
\end{equation}
where
$\widehat y_b^{\mathrm{env},m}
=
g_{\mathrm{spu}}^{m}
(\operatorname{Pool}_{\mathbf M}(\mathbf X_{\mathrm{spu}}^{m}))$
is produced by a linear classification head on the mask-averaged
context representation, $\ell_{\mathrm{env}}$ is the binary
cross-entropy over the two pseudo-environments, and $\varepsilon>0$
prevents division by zero.Importantly, $\mathcal L_{\mathrm{spu}}^{m}$ is activated only for
sports videos, whereas $\mathcal L_{\mathrm{ortho}}^{m}$ is applied to
all valid videos. Both terms share the single HAR-CDM coefficient
$\gamma_m$ in the total objective. The resulting task-oriented
representation $\mathbf H^{m}$ is passed to SGWIB, while
$\widetilde{\mathbf X}^{m}$ is retained as the detached source-side
geometry reference.
\subsection{Sliced Gromov--Monge Gap}
\label{subsec:sgmg}

SGWIB constructs a stochastic bottleneck from
$\mathbf H^{m}$ using Gaussian reparameterization, but unlike
conventional variational IB, no KL-based prior-matching penalty is
imposed. Instead, compression is regularized by SGMG, which measures
the excess relational distortion induced by the realized
source-to-bottleneck mapping relative to an optimal sliced structural
reference.

The stochastic channel scale and bottleneck representation are
\begin{equation}
\begin{aligned}
\mathbf s^{m}
&=
\boldsymbol{\mu}_{m}
+
\exp
\left(
\frac{1}{2}\boldsymbol{\ell}_{m}
\right)
\odot
\boldsymbol{\epsilon}^{m},
\qquad
\boldsymbol{\epsilon}^{m}
\sim
\mathcal N(\mathbf 0,\mathbf I),
\\
\mathbf z_{t}^{m}
&=
\mathbf h_{t}^{m}
\odot
\mathbf s^{m},
\qquad
t=1,\ldots,L,
\end{aligned}
\label{eq:channel_wise_stochastic_scaling}
\end{equation}
where $\boldsymbol{\mu}_{m}$ and $\boldsymbol{\ell}_{m}$ are learnable
channel-wise mean and log-variance parameters. One
$\mathbf s^{m}$ is sampled per video and shared across all temporal
positions, avoiding independently injected segment-wise perturbations.
The bottleneck therefore preserves the feature dimension and acts as a
multiplicative channel rather than a dimensionality-reducing encoder.
During inference, the deterministic mean scale
$\boldsymbol{\mu}_{m}$ is used by default.

Conditioned on a noise realization,
the bottleneck induces the deterministic mapping
\[
\phi_{\boldsymbol{\epsilon}^{m}}
\left(
\mathbf h_{t}^{m}
\right)
=
\mathbf h_{t}^{m}
\odot
\mathbf s^{m}.
\]
SGMG is evaluated for this realized mapping. Drawing one sample per
video in each forward pass yields a Monte Carlo approximation of its
expected stochastic objective.

\subsubsection{Geometry Construction and Sliced Projection}

For the current video,
\begin{equation}
\begin{aligned}
\mathcal I
&=
\left\{
t\in\{1,\ldots,L\}
\mid M_t=1
\right\},
\\
n
&=
|\mathcal I|.
\end{aligned}
\label{eq:sgmg_valid_indices}
\end{equation}

SGMG compares the bottleneck geometry against a detached source-side
reference:
\begin{equation}
\mathbf X_{\mathrm{geo}}^{m}
=
\operatorname{LN}_{0}
\left(
\operatorname{StopGrad}
\left(
\widetilde{\mathbf X}^{m}
\right)
\right),
\qquad
\mathbf Z_{\mathrm{geo}}^{m}
=
\operatorname{LN}_{0}
\left(
\mathbf Z^{m}
\right),
\label{eq:sgmg_geometry_normalization}
\end{equation}
where $\operatorname{LN}_{0}$ denotes affine-free layer
normalization. Stop-gradient fixes the source geometry while gradients
propagate through the bottleneck and its upstream mapping. Thus, the
implemented regularizer acts on the complete relational transformation
\[
\widetilde{\mathbf X}^{m}
\longrightarrow
\mathbf Z^{m},
\]
rather than only on
$\mathbf H^{m}\rightarrow\mathbf Z^{m}$.

For each valid segment $t\in\mathcal I$, the temporal coordinate is
rescaled over the valid range of the current video and mapped to
$\xi_t\in[-1,1]$, so that $\xi_t$ is defined only on $\mathcal I$ and
padded positions never enter the projection. The time-augmented source
and bottleneck features are then projected onto $P$ directions
$\boldsymbol{\theta}_{p}^{m}\in\mathbb R^{d_m+1}$ with
$\|\boldsymbol{\theta}_{p}^{m}\|_2=1$, drawn once at initialization
from the unit sphere $\mathbb S^{d_m}$ and held fixed throughout
training:
\begin{equation}
\begin{aligned}
u_{t}^{m,(p)}
&=
\left\langle
\boldsymbol{\theta}_{p}^{m},
\left[
\alpha_{\mathrm{time}}\xi_t;
\mathbf x_{\mathrm{geo},t}^{m}
\right]
\right\rangle,
\\
v_{t}^{m,(p)}
&=
\left\langle
\boldsymbol{\theta}_{p}^{m},
\left[
\alpha_{\mathrm{time}}\xi_t;
\mathbf z_{\mathrm{geo},t}^{m}
\right]
\right\rangle,
\end{aligned}
\quad
t\in\mathcal I,\;
p=1,\ldots,P.
\label{eq:sgmg_time_projection}
\end{equation}
The augmented vectors are $(d_m{+}1)$-dimensional, with
$\alpha_{\mathrm{time}}$ controlling the weight of the temporal
coordinate relative to the normalized features. The same projection
direction is used for source and bottleneck features so that each
one-dimensional comparison reflects structural distortion under a
shared view.

\subsubsection{Mapping Cost and Sliced Structural Reference}

For projection $p$, the mapping-induced relational cost is
\begin{equation}
\begin{aligned}
C_{\mathrm{map}}^{m,(p)}
&=
\\
&\frac{1}{n^{2}}
\sum_{s,t\in\mathcal I}
\left[
\left(
u_{s}^{m,(p)}
-
u_{t}^{m,(p)}
\right)^{2}
-
\left(
v_{s}^{m,(p)}
-
v_{t}^{m,(p)}
\right)^{2}
\right]^{2}.
\end{aligned}
\label{eq:sgmg_mapping_cost}
\end{equation}
The normalization is taken over all $n^{2}$ ordered pairs, including
the $n$ vanishing diagonal terms. This cost evaluates preservation of
pairwise relational organization under the prescribed temporal
correspondence, rather than pointwise feature reconstruction.

For the structural reference, projected source and bottleneck values
are sorted independently. Under equal cardinality, uniform empirical
weights, and squared-Euclidean relational costs, the
one-dimensional GW/GM optimum is attained by either the identity or
reversed correspondence:
\begin{equation}
C_{\mathrm{SGW}}^{m,(p)}
=
\min
\left\{
C_{\pi_{\mathrm{id}}}^{m,(p)},
C_{\pi_{\mathrm{rev}}}^{m,(p)}
\right\}.
\label{eq:sgmg_sliced_reference}
\end{equation}
Hence,
$C_{\mathrm{map}}^{m,(p)}$ measures the cost of the learned
correspondence, whereas
$C_{\mathrm{SGW}}^{m,(p)}$ provides the optimal
one-dimensional structural reference. Sorting is used only for this
reference computation and does not alter the temporal sequence used
for highlight prediction.

\subsubsection{SGMG Regularization}

The projection-level gap and video-level regularizer are
\begin{equation}
\begin{aligned}
G^{m,(p)}
&=
\left[
C_{\mathrm{map}}^{m,(p)}
-
C_{\mathrm{SGW}}^{m,(p)}
\right]_{+},
\\
\mathcal L_{\mathrm{SGMG,vid}}^{m}
&=
\frac{1}{P}
\sum_{p=1}^{P}
G^{m,(p)},
\end{aligned}
\label{eq:video_sgmg_regularization}
\end{equation}
where $[r]_{+}=\max(r,0)$.

Restoring the video index, the mini-batch objective is
\begin{equation}
\begin{aligned}
\mathcal L_{\mathrm{SGMG}}^{m}
&=
\frac{1}{|\mathcal B_{+}|}
\sum_{b\in\mathcal B_{+}}
\mathcal L_{\mathrm{SGMG,vid},b}^{m},
\\
\mathcal B_{+}
&=
\left\{
b\in\mathcal B
\mid n_b\geq1
\right\}.
\end{aligned}
\label{eq:sgmg_regularization}
\end{equation}
For $n_b=1$, the pairwise SGMG is defined as zero; videos with no valid
positions are excluded. Under the exact one-dimensional assumptions,
the gap is theoretically nonnegative, while the positive-part operator
is retained for numerical robustness.

A smaller SGMG indicates less excess relational distortion relative to
the optimal sliced structural reference. The validity mask
$\mathbf M_b$ determines the temporal positions participating in SGMG,
whereas the sports mask $s_b$ affects only pseudo-environment
supervision. Therefore, both sports and non-sports videos contribute
to SGMG whenever they contain valid temporal positions.

\subsubsection{Information-Theoretic Interpretation}
\label{subsubsec:sgmg_information_bound}

SGWIB follows the stochastic-encoding principle of Deep
VIB~\cite{alemi2017deep}, but replaces KL-based prior matching with
SGMG as a structure-aware dependence surrogate. For projection $p$ and
a fixed bottleneck-noise realization, define
\[
U^{m,(p)}
=
\left\{
u_t^{m,(p)}
\mid t\in\mathcal I
\right\},
\qquad
V^{m,(p)}
=
\left\{
v_t^{m,(p)}
\mid t\in\mathcal I
\right\}.
\]
Let
$\widehat I_{\mathrm{SKMI}}^{m}
(\boldsymbol{\epsilon}^{m})$
denote empirical kernelized dependence averaged over the $P$
projected source--bottleneck pairs.

\paragraph{Proposition 1.}
Conditioned on the realized upstream representation and a fixed
bottleneck-noise sample, assume equal cardinality, uniform empirical
weights, bounded normalized projected supports, and consistency
between the kernel and squared-Euclidean relational cost. Then,
\begin{equation}
\widehat I_{\mathrm{SKMI}}^{m}
\left(
\boldsymbol{\epsilon}^{m}
\right)
\leq
\frac{1}{2\sigma_{\kappa}^{2}}
\mathcal L_{\mathrm{SGMG,vid}}^{m}
\left(
\boldsymbol{\epsilon}^{m}
\right)
+
C_{\kappa,n}^{m},
\label{eq:conditional_sgmg_information_bound}
\end{equation}
where $\sigma_{\kappa}$ denotes the kernel bandwidth and
$C_{\kappa,n}^{m}=\frac{1}{P}\sum_{p=1}^{P}C_{\kappa,n}^{m,(p)}$ is
the projection-averaged Jensen-gap term, which remains finite under
the stated assumptions.

\paragraph{Proof Sketch.}
Conditioned on the upstream representation and
$\boldsymbol{\epsilon}^{m}$, the bottleneck becomes a deterministic
correspondence. Applying the empirical kernelized-dependence bound of
GWIB~\cite{yang2024gwib} to each projected pair gives
\begin{equation}
\begin{aligned}
\widehat I_{\kappa,n}^{m,(p)}
\left(
\boldsymbol{\epsilon}^{m}
\right)
&\leq
\frac{
C_{\mathrm{map}}^{m,(p)}
-
C_{\mathrm{GW}}^{m,(p)}
}{
2\sigma_{\kappa}^{2}
}
+
C_{\kappa,n}^{m,(p)}
\\
&=
\frac{
C_{\mathrm{map}}^{m,(p)}
-
C_{\mathrm{SGW}}^{m,(p)}
}{
2\sigma_{\kappa}^{2}
}
+
C_{\kappa,n}^{m,(p)}.
\end{aligned}
\label{eq:projection_sgmg_bound}
\end{equation}
Under one-dimensional squared-Euclidean costs, equal cardinality, and
uniform weights, the GW and Gromov--Monge optima coincide and are
attained by either the identity or reversed
correspondence~\cite{vayer2019sliced}; hence
$C_{\mathrm{GW}}^{m,(p)}=C_{\mathrm{SGW}}^{m,(p)}$.
Averaging over projection directions yields
Eq.~\eqref{eq:conditional_sgmg_information_bound}.

Affine-free layer normalization, bounded temporal coordinates, and
unit-norm projection directions ensure bounded projected empirical
supports. Taking expectation over bottleneck noise therefore preserves
the inequality:
\begin{equation}
\mathbb E_{\boldsymbol{\epsilon}^{m}}
\left[
\widehat I_{\mathrm{SKMI}}^{m}
\right]
\leq
\frac{1}{2\sigma_{\kappa}^{2}}
\mathbb E_{\boldsymbol{\epsilon}^{m}}
\left[
\mathcal L_{\mathrm{SGMG,vid}}^{m}
\right]
+
C_{\kappa,n}^{m}.
\label{eq:expected_sgmg_information_bound}
\end{equation}
Drawing one reparameterized noise sample per video and forward pass
provides a Monte Carlo estimate of this expected SGMG objective.

Proposition~1 does not upper-bound the full Shannon mutual information
$I(\widetilde{\mathbf X}^{m};\mathbf Z^{m})$. Instead, it shows that
expected SGMG controls an upper bound on empirical sliced kernelized
dependence, supporting its use as a tractable structure-aware
compression surrogate rather than an exact mutual-information
estimator.
\subsection{SGMG-Based Information Bottleneck for Video Highlight Detection}
\label{subsec:sgwib_vhd}
SGWIB follows the relevance--compression principle
\begin{equation}
\max\;
I
\left(
\mathbf Z^{m};\mathbf Y
\right)
-
\beta_m
I
\left(
\widetilde{\mathbf X}^{m};
\mathbf Z^{m}
\right),
\label{eq:conceptual_sgwib}
\end{equation}
where relevance is implemented through highlight regression and SGMG
serves as a structure-aware surrogate for compression. Unlike
KL-based variational IB, SGMG neither matches
$\mathbf Z^{m}$ to a predefined prior nor constitutes an exact
variational upper bound on Shannon mutual information; instead, it
regularizes excess relational distortion while controlling empirical
sliced kernelized dependence.

The modality-specific scorer $q_{\psi_m}$ predicts segment-level
highlight scores from $\mathbf Z^m$:
\begin{equation}
\begin{aligned}
\widehat{\mathbf y}^{m}
&=
q_{\psi_m}
\left(
\mathbf Z^{m}
\right),
\\
\mathcal L_{\mathrm{task}}^{m}
&=
\frac{
\displaystyle
\sum_{b\in\mathcal B}
\sum_{t=1}^{L}
M_{b,t}
\left(
\widehat y_{b,t}^{m}
-
y_{b,t}
\right)^2
}{
\displaystyle
\sum_{b\in\mathcal B}
\sum_{t=1}^{L}
M_{b,t}
+
\varepsilon
}.
\end{aligned}
\label{eq:highlight_prediction_and_task_loss}
\end{equation}
Note that the scorer takes $\mathbf Z^{m}$ as its input, whereas SGMG
contributes only a regularization term and does not lie on the
forward path.

The complete modality-specific objective combines the task loss with
two independently weighted regularizers:
\begin{equation}
\begin{aligned}
\mathcal L_{\mathrm{total}}^{m}
={}&
\mathcal L_{\mathrm{task}}^{m}
+
\beta_m \mathcal L_{\mathrm{SGMG}}^{m}
\\
&+
\gamma_m
\bigl(
\mathcal L_{\mathrm{ortho}}^{m}
+
\mathcal L_{\mathrm{spu}}^{m}
\bigr),
\qquad
m\in\mathcal M=\{v,a\}.
\end{aligned}
\label{eq:total_training_loss}
\end{equation}
Here $\beta_m$ controls SGMG-based structural compression and
$\gamma_m$ scales the two HAR-CDM terms jointly. Although the
orthogonality constraint applies to every valid video while the
pseudo-environment supervision is restricted to the sports subset,
the two are tied to a single coefficient, so HAR-CDM contributes one
tunable weight rather than two. Visual and audio branches use separate
parameters and are trained independently, without cross-modal
alignment or fusion.
\section{Experiments}
\label{sec:exp}

\subsection{Experimental Setup}
\label{subsec:experimental_setup}

\subsubsection{Datasets}
\label{subsubsec:datasets}

We evaluate SGWIB on MrHiSum and MoSu, two large-scale video
highlight detection datasets derived from YouTube-8M
~\cite{abu2016youtube} and annotated using YouTube Most Replayed
statistics. MrHiSum~\cite{sul2023mr} contains 31,892 videos from
3,509 categories, retaining videos with more than 50,000 views to
improve annotation reliability. MoSu~\cite{kim2026triplesumm}
contains 52,678 videos from 3,406 categories, corresponding to
approximately 4,000 hours of content with an average duration of
272.3 seconds. MoSu provides visual, audio, and textual information,
while textual features are not used in this work.

The visual branch serves as the primary experimental setting for
comparison with existing VHD methods. The same single-modal
formulation is independently applied to audio as auxiliary
modality-general validation, without visual--audio fusion or
cross-modal interaction. For MrHiSum, temporally aligned audio
features are extracted from the corresponding source videos.
Table~\ref{tab:dataset_summary} summarizes the dataset configurations.

\begin{table}[!t]
\centering
\caption{Summary of the datasets used in the experiments.}
\label{tab:dataset_summary}
\small
\renewcommand{\arraystretch}{1.10}
\setlength{\tabcolsep}{6pt}
\begin{tabular}{@{}lrrl@{}}
\toprule
\textbf{Dataset}
& \textbf{Videos}
& \textbf{Categories}
& \textbf{Train/Val/Test} \\
\midrule
MrHiSum
& 31,892
& 3,509
& $80\%/10\%/10\%$ \\
MoSu
& 52,678
& 3,406
& $60\%/20\%/20\%$ \\
\bottomrule
\end{tabular}
\end{table}

\subsubsection{Implementation Details}
\label{subsubsec:implementation_details}
The modality-specific backbone $E_m$ is instantiated with
CSTA~\cite{son2024csta}, so that the reported gains are attributable to
the proposed regularizers rather than to a new feature extractor. Both
datasets use frame-level YouTube-8M features ($d_v=1024$, $d_a=128$),
and sports videos are identified by the sports-related category labels
of the YouTube-8M taxonomy rather than by manual re-annotation, giving
$2{,}947$ of $27{,}892$ MrHiSum training videos ($10.57\%$) and
$3{,}406$ of $42{,}152$ MoSu training videos ($8.08\%$). The scorer is
an MLP $d_m\!\rightarrow\!512\!\rightarrow\!256\!\rightarrow\!1$ with
GELU, layer normalization after the first hidden layer, dropout $0.1$,
and a sigmoid output. HAR-CDM uses 8 attention heads with dropout
$0.1$. For SGMG we set $P=128$ and $\alpha_{\mathrm{time}}=0.6$, and
initialize $\boldsymbol\mu_m$ and $\boldsymbol\ell_m$ to $1.0$ and
$-9.0$. Our implementation also retains a gradient-reversal branch on
the task-oriented pathway, but its weight is zero in all reported
experiments.

All experiments run on a single NVIDIA A100 GPU. Each modality-specific
model is independently trained for 50 epochs with a batch size of 32 and
an initial learning rate of $1\times10^{-4}$, with hyperparameters
selected on the validation set and the test set reserved for final
evaluation. $\beta_m$ and $\gamma_m$ are selected on the validation set
over a log grid spanning $10^{-10}$ to $10^{-3}$ and the set
$\{5{\times}10^{-4},10^{-3},5{\times}10^{-3},10^{-2},
5{\times}10^{-2},10^{-1},5{\times}10^{-1},1\}$, respectively, and are
varied separately for sensitivity analysis with the other coefficient
held at its selected value. The structural-preservation analysis
additionally uses $\beta_v=10^{-1}$ as a strong-compression stress
test.
\subsubsection{Evaluation Metrics}
\label{subsubsec:evaluation_metrics}

Following the rank-based video-summarization protocol
~\cite{otani2019rethinking}, temporal importance prediction is
evaluated using Kendall's $\tau$ and Spearman's $\rho$, which measure
pairwise ranking concordance and monotonic association between the
predicted and ground-truth rankings, respectively
~\cite{kendall1945treatment,spearman1904proof}. For highlight
retrieval, we follow Mr.~HiSum~\cite{sul2023mr}: each video is divided
into non-overlapping five-second segments, and predicted and
ground-truth scores are averaged within each segment. For
$\mathrm{mAP}@q$, the top $q\%$ of segments according to ground-truth
importance are treated as positives, while all segments are ranked by
their predicted scores. Average precision is computed per video and
then averaged over the test set. We report
$q\in\{15,30,50\}$, where $15\%$ and $50\%$ follow the original
protocol and $30\%$ provides an intermediate highlight budget.
Higher values indicate better performance for all metrics.

\subsection{Comparison with State-of-the-Art Methods}
\label{subsec:comparison_existing_methods}

We compare the visual instantiation of SGWIB with representative video
summarization and VHD methods, including
VASNet~\cite{fajtl2019summarizing},
PGL-SUM~\cite{apostolidis2021combining},
iPTNet~\cite{jiang2022joint},
the SL-module~\cite{xu2021crosscategory},
CSTA~\cite{son2024csta},
SUM-GAN~\cite{mahasseni2017unsupervised},
AC-SUM-GAN~\cite{apostolidis2021ac}, and
SummDiff~\cite{kim2025summdiff}. To ensure a consistent input setting,
the main comparison is restricted to methods using visual features at
inference; the independently trained audio branch is evaluated
separately in the subsequent diagnostic analyses.

\begin{table*}[t]
\centering
\caption{Visual-only comparison with state-of-the-art methods on
MrHiSum and MoSu. Best and second-best results within each dataset are
highlighted in bold and underlined, respectively.}
\label{tab:sota_visual_comparison}

\normalsize

\renewcommand{\arraystretch}{1.08}
\setlength{\tabcolsep}{2.2pt}

\begin{tabular}{llccccc}
\toprule
\textbf{Dataset}
& \textbf{Method}
& \boldmath$\tau\uparrow$
& \boldmath$\rho\uparrow$
& \textbf{mAP@50}\boldmath$\uparrow$
& \textbf{mAP@30}\boldmath$\uparrow$
& \textbf{mAP@15}\boldmath$\uparrow$ \\
\midrule

\multirow{9}{*}{MrHiSum}
& VASNet~\cite{fajtl2019summarizing}
& $0.068$ & $0.100$ & $58.67$ & $39.67$ & $25.21$ \\

& PGL-SUM~\cite{apostolidis2021combining}
& $0.097$ & $0.141$ & $53.77$ & $35.25$ & $21.52$ \\

& iPTNet~\cite{jiang2022joint}
& $0.020$ & $0.029$ & $55.53$ & -- & $22.74$ \\

& SL-module~\cite{xu2021crosscategory}
& $0.062$ & $0.091$ & $58.27$ & $38.88$ & $24.44$ \\

& SUM-GAN~\cite{mahasseni2017unsupervised}
& $0.067$ & $0.095$ & $56.62$ & -- & $23.56$ \\

& AC-SUM-GAN~\cite{apostolidis2021ac}
& $0.012$ & $0.018$ & $55.35$ & -- & $21.88$ \\

& CSTA~\cite{son2024csta}
& $0.157$ & $0.210$ & $63.34$ & $45.87$ & $28.95$ \\

& SummDiff~\cite{kim2025summdiff}
& $\underline{0.161}$
& $\underline{0.224}$
& $\underline{65.24}$
& $\underline{46.47}$
& $\mathbf{33.31}$ \\

& \textbf{SGWIB (Ours)}
& $\mathbf{0.192}$
& $\mathbf{0.255}$
& $\mathbf{66.11}$
& $\mathbf{47.22}$
& $\underline{30.08}$ \\

\midrule

\multirow{6}{*}{MoSu}
& VASNet~\cite{fajtl2019summarizing}
& $0.143$ & $0.207$ & $62.82$ & $43.75$ & $28.50$ \\

& PGL-SUM~\cite{apostolidis2021combining}
& $0.163$ & $0.234$ & $64.91$ & $46.10$ & $29.80$ \\

& SL-module~\cite{xu2021crosscategory}
& $0.144$ & $0.208$ & $62.92$ & $44.00$ & $28.67$ \\

& CSTA~\cite{son2024csta}
& $\underline{0.240}$
& $\underline{0.318}$
& $\underline{68.33}$
& $\underline{50.23}$
& $\mathbf{32.93}$ \\

& SummDiff~\cite{kim2025summdiff}
& $0.171$ & $0.245$ & $65.76$ & $46.22$ & $29.96$ \\

& \textbf{SGWIB (Ours)}
& $\mathbf{0.249}$
& $\mathbf{0.331}$
& $\mathbf{68.67}$
& $\mathbf{50.35}$
& $\underline{32.65}$ \\

\bottomrule
\end{tabular}
\end{table*}

As shown in Table~\ref{tab:sota_visual_comparison}, SGWIB achieves the
best Kendall's $\tau$, Spearman's $\rho$, $\mathrm{mAP}@50$, and
$\mathrm{mAP}@30$ on both datasets, and is the only method ranked
first on these four metrics across both benchmarks. On MrHiSum,
SGWIB reaches $0.192$, $0.255$, $66.11$, and $47.22$, improving over
the second-best SummDiff by $0.031$, $0.031$, $0.87$, and $0.75$,
respectively. Relative to the CSTA backbone, the gains are $0.035$ in
$\tau$ and $0.045$ in $\rho$, corresponding to relative improvements
of over $20\%$, together with consistent gains at all three retrieval
budgets, including $+1.13$ on $\mathrm{mAP}@15$. On MoSu, SGWIB
attains $0.248$, $0.329$, $68.58$, and $50.29$, exceeding CSTA by
$0.008$, $0.011$, $0.25$, and $0.06$. The smaller margin on MoSu is
consistent with the substantially stronger backbone performance on
this dataset ($\tau=0.240$ versus $0.157$ on MrHiSum), which leaves
less headroom for further improvement.

Two additional observations are worth noting. First, the improvements
are more pronounced on the rank-based metrics than on the retrieval
metrics, and within the latter they decrease as the highlight budget
narrows. This pattern is consistent with the intended effect of SGMG:
preserving inter-segment relational structure primarily benefits the
global ordering of temporal importance, whereas identifying a very
small set of top-ranked segments depends more on local
discriminability. Accordingly, SummDiff attains the highest
$\mathrm{mAP}@15$ on MrHiSum, and the CSTA backbone remains slightly
better on MoSu ($32.93$ versus $32.65$). Second, method rankings are
not stable across the two benchmarks: SummDiff is the second-best
method on MrHiSum but falls below both CSTA and PGL-SUM on MoSu,
whereas SGWIB ranks first on $\tau$, $\rho$, $\mathrm{mAP}@50$, and
$\mathrm{mAP}@30$ on both. This suggests that structure-aware
regularization transfers more reliably across data distributions than
the compared alternatives.

\subsection{Qualitative Results}
\label{subsec:qualitative_results}
\begin{figure}[!t]
    \centering
    \subfloat[MrHiSum.\label{fig:qualitative_mrhisum}]{
        \includegraphics[width=0.9\columnwidth]
        {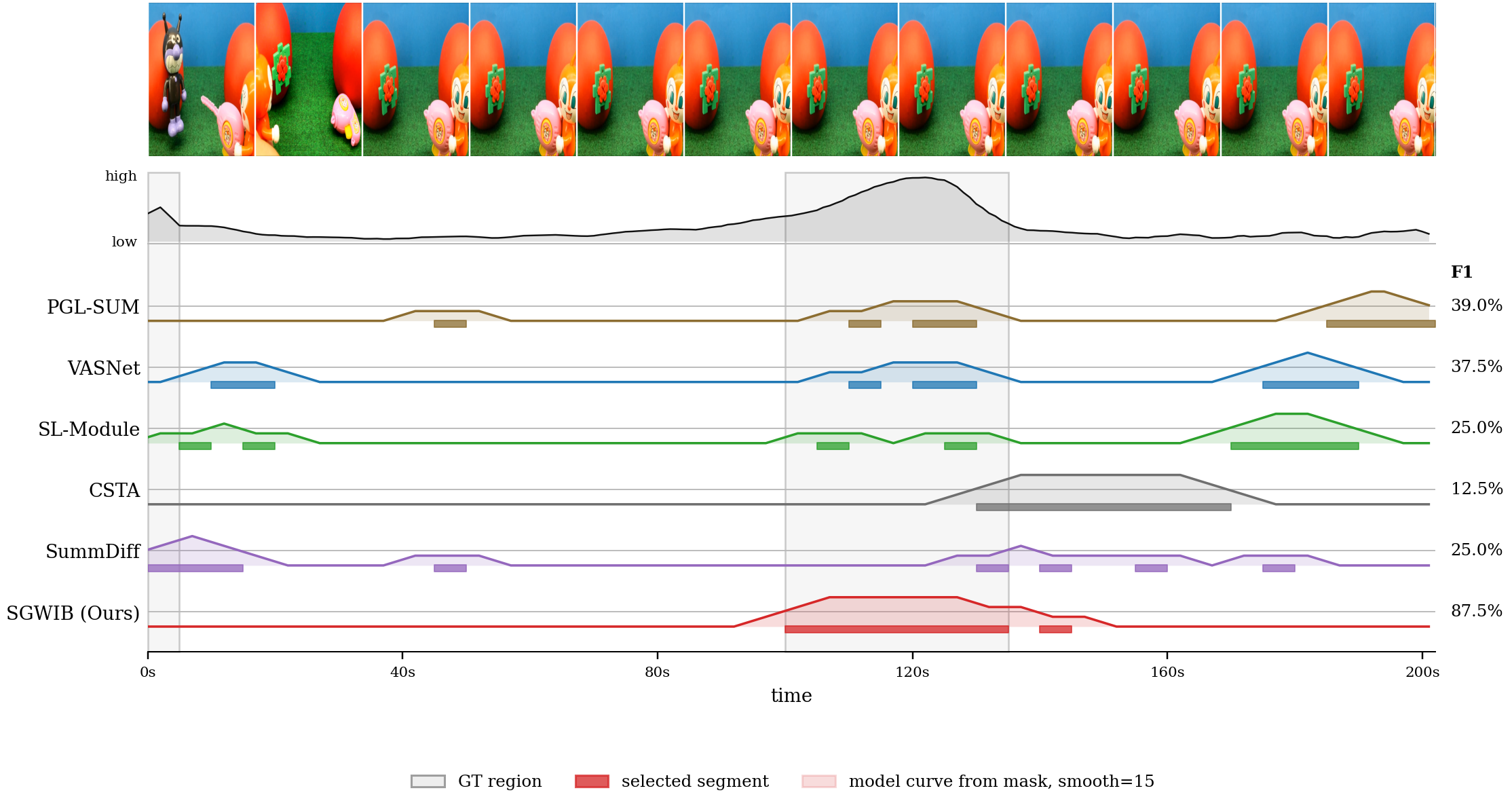}
    }
    \vspace{0.5em}
    \subfloat[MoSu.\label{fig:qualitative_mosu}]{
        \includegraphics[width=0.9\columnwidth]
        {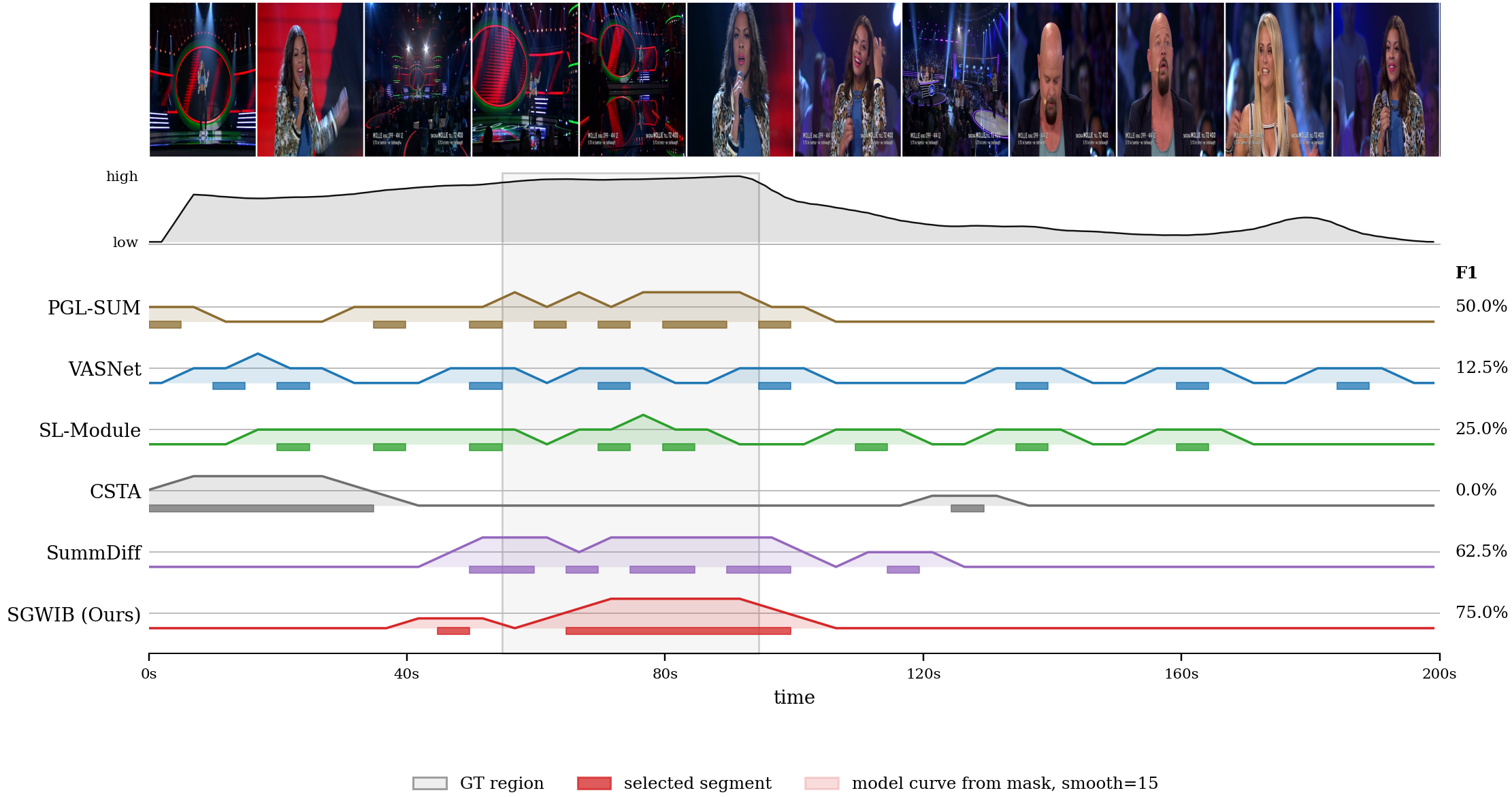}
    }
    \vspace{0.5em}
    \subfloat[YouTube video.\label{fig:qualitative_youtube}]{
        \includegraphics[width=0.9\columnwidth]
        {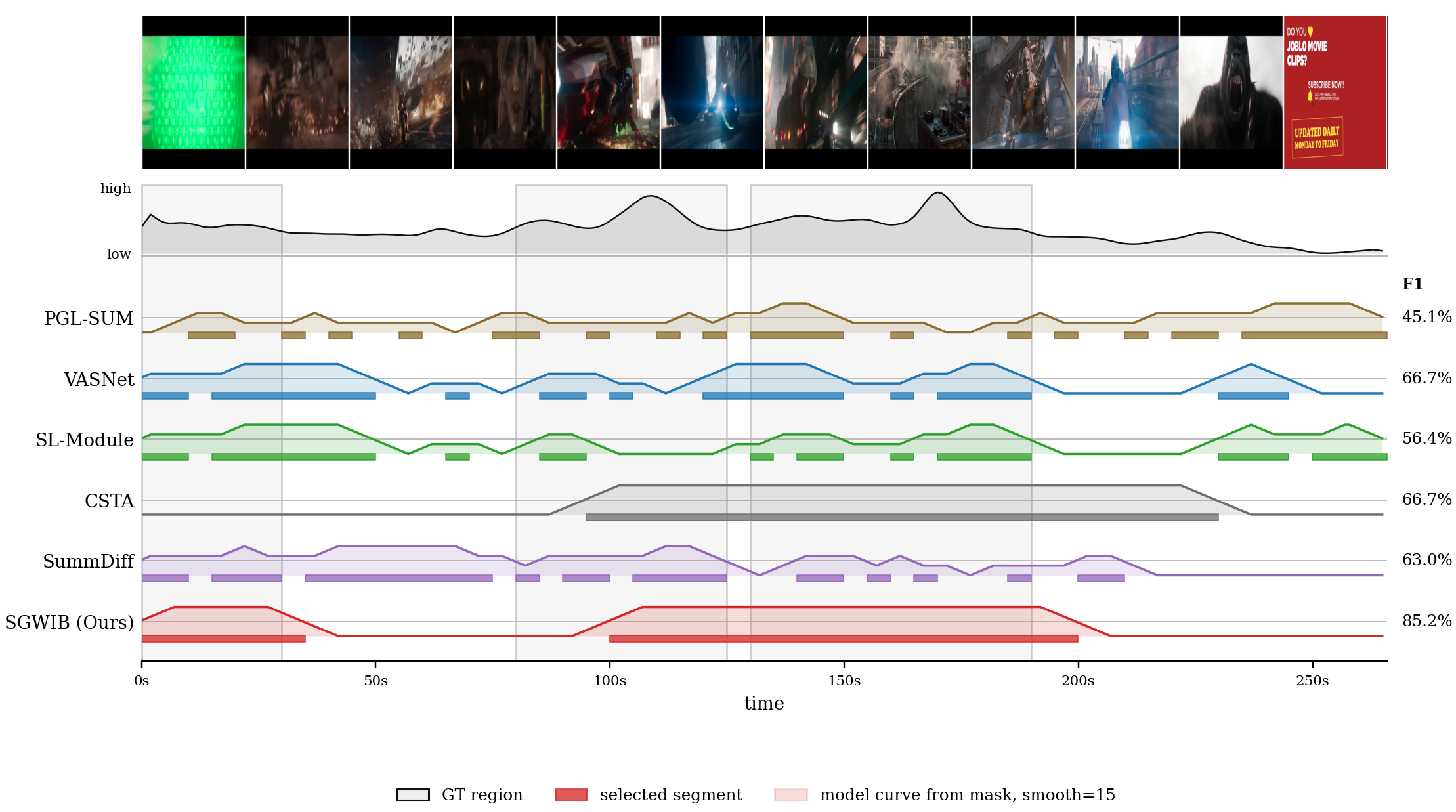}
    }
    \caption{Qualitative comparison of temporal highlight prediction
    on (a) MrHiSum, (b) MoSu, and (c) an additional YouTube video.
    The ground-truth highlight intervals of the YouTube example are
    manually annotated for qualitative evaluation.}
    \label{fig:qualitative_comparison}
\end{figure}

Figure~\ref{fig:qualitative_comparison} compares temporal highlight
predictions on MrHiSum, MoSu, and an additional YouTube video. Gray
regions denote ground-truth highlight intervals, colored curves denote
predicted temporal importance scores, and horizontal bars mark the
selected highlight segments. Overall, SGWIB produces responses more
concentrated within the annotated highlight regions, whereas several
competing methods exhibit more dispersed activations or select
additional segments outside the target intervals.

On MrHiSum, the dominant response of SGWIB closely overlaps the main
ground-truth interval, yielding an F1 score of $87.5\%$. On MoSu, it
focuses mainly on the central highlight region and achieves $75.0\%$
F1. To examine behavior beyond the benchmark datasets, we additionally
evaluate a YouTube video with manually annotated highlight intervals.
SGWIB captures multiple temporally separated highlight regions and
achieves $85.2\%$ F1. These examples complement the quantitative
evaluation by showing more temporally localized predictions with
stronger correspondence to the annotated highlight structure.
\subsection{Ablation and Diagnostic Analysis}
\label{subsec:ablation_diagnostic}
We further examine SGWIB from five perspectives: the contribution of
individual components, alternative information-bottleneck formulations,
computational efficiency, structural preservation, and hyperparameter
sensitivity. The visual branch serves as the primary setting throughout.
For the component, bottleneck-comparison, and sensitivity analyses, we
additionally report independently trained audio models to verify that
the observed behavior generalizes across feature modalities.

\subsubsection{Component Ablation}
\label{subsubsec:component_ablation}

We compare four configurations: the backbone alone, the backbone with
HAR-CDM, the backbone with SGWIB, and the full model. Removing HAR-CDM
disables task/context decomposition together with its orthogonality
regularization and sports-specific pseudo-environment supervision;
removing SGWIB disables the SGMG-based bottleneck regularization. For
every configuration, $\gamma_m$ and $\beta_m$ are selected on the
validation set.

\begin{table*}[!t]
\centering
\caption{Component ablation of HAR-CDM and SGWIB under the best
validation-selected hyperparameters. For configurations that include
HAR-CDM or SGWIB, the selected $\gamma_m$ and $\beta_m$ are reported.
The best and second-best results within each dataset and modality are
shown in bold and underlined, respectively.}
\label{tab:component_ablation}

\renewcommand{\arraystretch}{1.08}
\setlength{\tabcolsep}{2.2pt}

\begin{tabular}{@{}llccccccccc@{}}
\toprule
\textbf{Dataset}
& \textbf{Modality}
& \textbf{HAR-CDM}
& \textbf{SGWIB}
& \boldmath$\gamma_m$
& \boldmath$\beta_m$
& \boldmath$\tau\uparrow$
& \boldmath$\rho\uparrow$
& \textbf{mAP@50}\boldmath$\uparrow$
& \textbf{mAP@30}\boldmath$\uparrow$
& \textbf{mAP@15}\boldmath$\uparrow$ \\
\midrule

\multirow{8}{*}{MrHiSum}
& \multirow{4}{*}{Visual}
& \xmark & \xmark & -- & --
& $0.157$ & $0.210$ & $64.15$ & $45.87$ & $28.95$ \\

&
& \cmark & \xmark & $0.05$ & --
& $0.180$ & $\underline{0.242}$
& $65.38$ & $45.88$ & $28.30$ \\

&
& \xmark & \cmark & -- & $10^{-8}$
& $\underline{0.181}$ & $\underline{0.242}$
& $\underline{65.66}$ & $\underline{46.79}$
& $\underline{29.40}$ \\

&
& \cmark & \cmark & $0.01$ & $10^{-6}$
& $\mathbf{0.192}$ & $\mathbf{0.255}$
& $\mathbf{66.11}$ & $\mathbf{47.22}$
& $\mathbf{30.08}$ \\

\cmidrule(lr){2-11}

&
\multirow{4}{*}{Audio}
& \xmark & \xmark & -- & --
& $0.141$ & $0.191$ & $63.39$ & $45.87$ & $28.88$ \\

&
& \cmark & \xmark & $0.01$ & --
& $\underline{0.173}$ & $\underline{0.230}$
& $\underline{65.22}$ & $\mathbf{46.86}$
& $\underline{29.74}$ \\

&
& \xmark & \cmark & -- & $10^{-9}$
& $0.169$ & $0.224$
& $64.90$ & $45.86$ & $28.72$ \\

&
& \cmark & \cmark & $0.01$ & $10^{-8}$
& $\mathbf{0.179}$ & $\mathbf{0.239}$
& $\mathbf{65.45}$ & $\underline{46.52}$
& $\mathbf{29.98}$ \\

\midrule

\multirow{8}{*}{MoSu}
& \multirow{4}{*}{Visual}
& \xmark & \xmark & -- & --
& $0.240$ & $0.318$
& $68.33$ & $50.23$ & $\mathbf{32.93}$ \\

&
& \cmark & \xmark & $0.01$ & --
& $\underline{0.245}$ & $\underline{0.327}$
& $\underline{68.49}$ & $\underline{50.30}$
& $\underline{32.88}$ \\

&
& \xmark & \cmark & -- & $10^{-7}$
& $0.244$ & $0.322$
& $68.40$ & $\mathbf{50.32}$
& $32.75$ \\

&
& \cmark & \cmark & $0.01$ & $10^{-5}$
& $\mathbf{0.249}$ & $\mathbf{0.331}$
& $\mathbf{68.67}$ & $50.35$
& $32.65$ \\

\cmidrule(lr){2-11}

&
\multirow{4}{*}{Audio}
& \xmark & \xmark & -- & --
& $0.200$ & $0.267$
& $65.82$ & $47.11$ & $\underline{29.93}$ \\

&
& \cmark & \xmark & $0.5$ & --
& $\underline{0.209}$ & $\underline{0.282}$
& $\underline{66.42}$ & $\underline{47.27}$
& $29.78$ \\

&
& \xmark & \cmark & -- & $10^{-4}$
& $0.207$ & $0.277$
& $66.22$ & $47.17$
& $29.82$ \\

&
& \cmark & \cmark & $0.5$ & $10^{-3}$
& $\mathbf{0.215}$ & $\mathbf{0.288}$
& $\mathbf{66.50}$ & $\mathbf{47.37}$
& $\mathbf{30.04}$ \\

\bottomrule
\end{tabular}
\end{table*}

As shown in Table~\ref{tab:component_ablation}, HAR-CDM and SGWIB each
improve over the backbone in most settings, and combining them yields
the strongest overall performance. On MrHiSum Visual, the full model
raises Kendall's $\tau$ from $0.157$ to $0.192$ and Spearman's $\rho$
from $0.210$ to $0.255$, and improves $\mathrm{mAP}@50$,
$\mathrm{mAP}@30$, and $\mathrm{mAP}@15$ from $64.15$, $45.87$, and
$28.95$ to $66.11$, $47.22$, and $30.08$, respectively. On MoSu Visual,
the full model again attains the best $\tau$, $\rho$, and
$\mathrm{mAP}@50$, whereas the backbone remains marginally stronger on
$\mathrm{mAP}@15$. The independently trained audio models follow the
same overall trend. These results are consistent with the two modules
playing complementary roles: HAR-CDM encourages task/context separation
through orthogonality regularization and sports-specific
pseudo-environment supervision, whereas SGWIB regularizes the structural
compression of the task-oriented representation across all videos,
without requiring domain-specific annotation.
\subsubsection{Comparison of Information Bottleneck Formulations}
\label{subsubsec:ib_formulation_comparison}

We compare conventional KLIB, full GWIB, and SGWIB while keeping the
backbone, HAR-CDM, prediction head, regression objective, data splits,
and optimization settings unchanged wherever applicable. KLIB
regularizes the latent distribution toward a predefined prior, GWIB
models relational structure through full Gromov--Wasserstein
optimization, and SGWIB uses SGMG to quantify excess relational
distortion induced by the learned source-to-bottleneck correspondence.
KLIB and SGWIB are trained for the full 50 epochs. Because full GW
optimization is substantially more expensive, GWIB is included only
in the efficiency analysis; its limited-epoch prediction results are
not directly comparable.

\begin{table*}[!t]
\centering
\caption{Task-level comparison of KLIB and SGWIB. All network
components and training settings are identical except for the
information-bottleneck regularizer. The optimal $\beta$ is selected
independently on the validation set.}
\label{tab:kl_sgw_ib_comparison}

\renewcommand{\arraystretch}{1.08}
\setlength{\tabcolsep}{4pt}

\begin{tabular}{@{}lllcccccc@{}}
\toprule
\textbf{Dataset}
& \textbf{Modality}
& \textbf{IB Formulation}
& \textbf{Optimal \boldmath$\beta$}
& \boldmath$\tau\uparrow$
& \boldmath$\rho\uparrow$
& \textbf{mAP@50}\boldmath$\uparrow$
& \textbf{mAP@30}\boldmath$\uparrow$
& \textbf{mAP@15}\boldmath$\uparrow$ \\
\midrule

\multirow{4}{*}{MrHiSum}
& \multirow{2}{*}{Visual}
& KLIB
& $10^{-8}$
& $0.181$
& $0.242$
& $65.52$
& $47.02$
& $29.67$ \\

&
& SGWIB
& $10^{-6}$
& $\mathbf{0.192}$
& $\mathbf{0.255}$
& $\mathbf{66.11}$
& $\mathbf{47.22}$
& $\mathbf{30.08}$ \\

\cmidrule(lr){2-9}

&
\multirow{2}{*}{Audio}
& KLIB
& $10^{-8}$
& $0.166$
& $0.220$
& $64.78$
& $46.05$
& $28.99$ \\

&
& SGWIB
& $10^{-8}$
& $\mathbf{0.179}$
& $\mathbf{0.239}$
& $\mathbf{65.45}$
& $\mathbf{46.52}$
& $\mathbf{29.98}$ \\

\midrule

\multirow{4}{*}{MoSu}
& \multirow{2}{*}{Visual}
& KLIB
& $10^{-3}$
& $0.244$
& $0.326$
& $68.40$
& $50.03$
& $32.21$ \\

&
& SGWIB
& $10^{-5}$
& $\mathbf{0.249}$
& $\mathbf{0.331}$
& $\mathbf{68.67}$
& $\mathbf{50.35}$
& $\mathbf{32.65}$ \\

\cmidrule(lr){2-9}

&
\multirow{2}{*}{Audio}
& KLIB
& $10^{-6}$
& $0.203$
& $0.272$
& $66.07$
& $47.21$
& $29.99$ \\

&
& SGWIB
& $10^{-3}$
& $\mathbf{0.215}$
& $\mathbf{0.288}$
& $\mathbf{66.50}$
& $\mathbf{47.37}$
& $\mathbf{30.04}$ \\

\bottomrule
\end{tabular}
\end{table*}
Table~\ref{tab:kl_sgw_ib_comparison} shows that SGWIB outperforms KLIB
across both datasets and modalities. On the primary visual branch,
SGWIB improves all five metrics on MrHiSum and MoSu. The audio branch
shows the same trend, with Kendall's $\tau$ and Spearman's $\rho$
improvements of $0.013$ and $0.019$ on MrHiSum and $0.012$ and
$0.016$ on MoSu, respectively. These results indicate that replacing
KL-based prior matching with SGMG yields more favorable task-level
behavior across feature modalities.

\subsubsection{Computational Efficiency Analysis}
\label{subsubsec:computational_efficiency}
Using the MrHiSum visual branch, we compare KLIB, SGWIB, and GWIB under
identical hardware, batch size, data split, and optimization settings.
Each epoch contains $1{,}744$ iterations, and the reported time and
throughput are averaged over 10 consecutive epochs to reduce runtime
fluctuation.

\begin{table}[!t]
\centering
\caption{Average per-epoch training cost on the MrHiSum visual branch.
Each epoch contains $1{,}744$ iterations, and results are averaged over
10 epochs. Relative cost is normalized to KLIB.}
\label{tab:ib_training_cost}
\fontsize{9.8}{10.6}\selectfont
\renewcommand{\arraystretch}{1.05}
\setlength{\tabcolsep}{3.2pt}
\begin{tabular}{lcccc}
\toprule
\textbf{Method}
& \textbf{Time}
& \textbf{Iters.}
& \textbf{it/s}
& \textbf{Cost} \\
\midrule
KLIB
& 18m 40s
& $1{,}744$
& 1.56
& $1.00\times$ \\
SGWIB
& 20m 34s
& $1{,}744$
& 1.41
& $1.10\times$ \\
GWIB
& 2h 15m 58s
& $1{,}744$
& 0.21
& $7.28\times$ \\
\bottomrule
\end{tabular}
\end{table}

As shown in Table~\ref{tab:ib_training_cost}, SGWIB takes 20m 34s per
epoch against 18m 40s for KLIB, an overhead of only $10.2\%$. Full GWIB
instead requires 2h 15m 58s, which is $7.28\times$ the cost of KLIB and
$6.61\times$ that of SGWIB, and its throughput drops from 1.56 to 0.21
iterations per second. SGWIB thus retains nearly all of the efficiency
of KLIB while providing structure-aware regularization, whereas full
Gromov--Wasserstein optimization scales poorly on long video sequences.
This gap also explains why GWIB is excluded from the task-level
comparison in Section~\ref{subsubsec:ib_formulation_comparison}: running
the full 50-epoch schedule for every dataset and modality is
prohibitively expensive.
\subsubsection{Structural Preservation Analysis}
\label{subsubsec:structural_preservation}

We use the primary visual branch to examine how KLIB and SGWIB affect
temporal relational structure. The analysis uses a representative
MrHiSum video with $n=98$ valid segments. Let $\widetilde{\mathbf X}^{v}$
and $\mathbf Z^{v}$ denote the input sequence and the bottleneck
representation, so that the diagnostic measures the same
transformation that SGMG regularizes during training. Their
segment-level vectors are concatenated and jointly standardized along
each feature dimension, yielding
$\widetilde{\mathbf x}_{t}^{v,\mathrm{std}}$ and
$\mathbf z_{t}^{v,\mathrm{std}}$.

Structural preservation is quantified by pairwise distortion, paired
shift, and centroid shift:
\begin{equation}
\begin{aligned}
D_{s,t}^{v,\mathrm{pre}}
&=
\left\|
\widetilde{\mathbf x}_{s}^{v,\mathrm{std}}
-
\widetilde{\mathbf x}_{t}^{v,\mathrm{std}}
\right\|_2,
\\
D_{s,t}^{v,\mathrm{post}}
&=
\left\|
\mathbf z_{s}^{v,\mathrm{std}}
-
\mathbf z_{t}^{v,\mathrm{std}}
\right\|_2,
\\
d_{\max}^{v}
&=
\max\!\left\{
\max_{s,t\in\mathcal I}
D_{s,t}^{v,\mathrm{pre}},
\;
\max_{s,t\in\mathcal I}
D_{s,t}^{v,\mathrm{post}}
\right\},
\\
\mathcal D_{\mathrm{pair}}^{v}
&=
\frac{1}{n^{2}}
\sum_{s,t\in\mathcal I}
\left|
\frac{
D_{s,t}^{v,\mathrm{pre}}
-
D_{s,t}^{v,\mathrm{post}}
}{
d_{\max}^{v}+\varepsilon
}
\right|,
\\
\mathcal D_{\mathrm{paired}}^{v}
&=
\frac{1}{n}
\sum_{t\in\mathcal I}
\left\|
\widetilde{\mathbf x}_{t}^{v,\mathrm{std}}
-
\mathbf z_{t}^{v,\mathrm{std}}
\right\|_2,
\\
\mathcal D_{\mathrm{centroid}}^{v}
&=
\left\|
\frac{1}{n}
\sum_{t\in\mathcal I}
\widetilde{\mathbf x}_{t}^{v,\mathrm{std}}
-
\frac{1}{n}
\sum_{t\in\mathcal I}
\mathbf z_{t}^{v,\mathrm{std}}
\right\|_2.
\end{aligned}
\label{eq:structural_preservation_metrics}
\end{equation}
$\mathcal D_{\mathrm{pair}}^{v}$ measures changes in normalized
pairwise geometry, $\mathcal D_{\mathrm{paired}}^{v}$ the average
displacement of corresponding segments, and
$\mathcal D_{\mathrm{centroid}}^{v}$ the shift of the representation
center; lower values indicate better preservation. All three are
computed in the standardized feature space, and the normalization of
$\mathcal D_{\mathrm{pair}}^{v}$ runs over all $n^{2}$ ordered pairs.
Unlike the training objective, these diagnostics compare the two point
sets directly rather than through one-dimensional slices, so they
provide an independent view of the same structural change.
For visualization, the same representations are additionally embedded
into a shared two-dimensional space using t-SNE with PCA
initialization, perplexity 30, automatic learning-rate selection, and
random seed 42; since t-SNE does not strictly preserve global
distances, it serves only as a qualitative complement.

\begin{figure*}[!t]
    \centering
    \subfloat[
        Pairwise distortion
        $\mathcal D_{\mathrm{pair}}^{v}$.
        \label{fig:pairwise_distortion}
    ]{
        \includegraphics[width=0.31\textwidth]
        {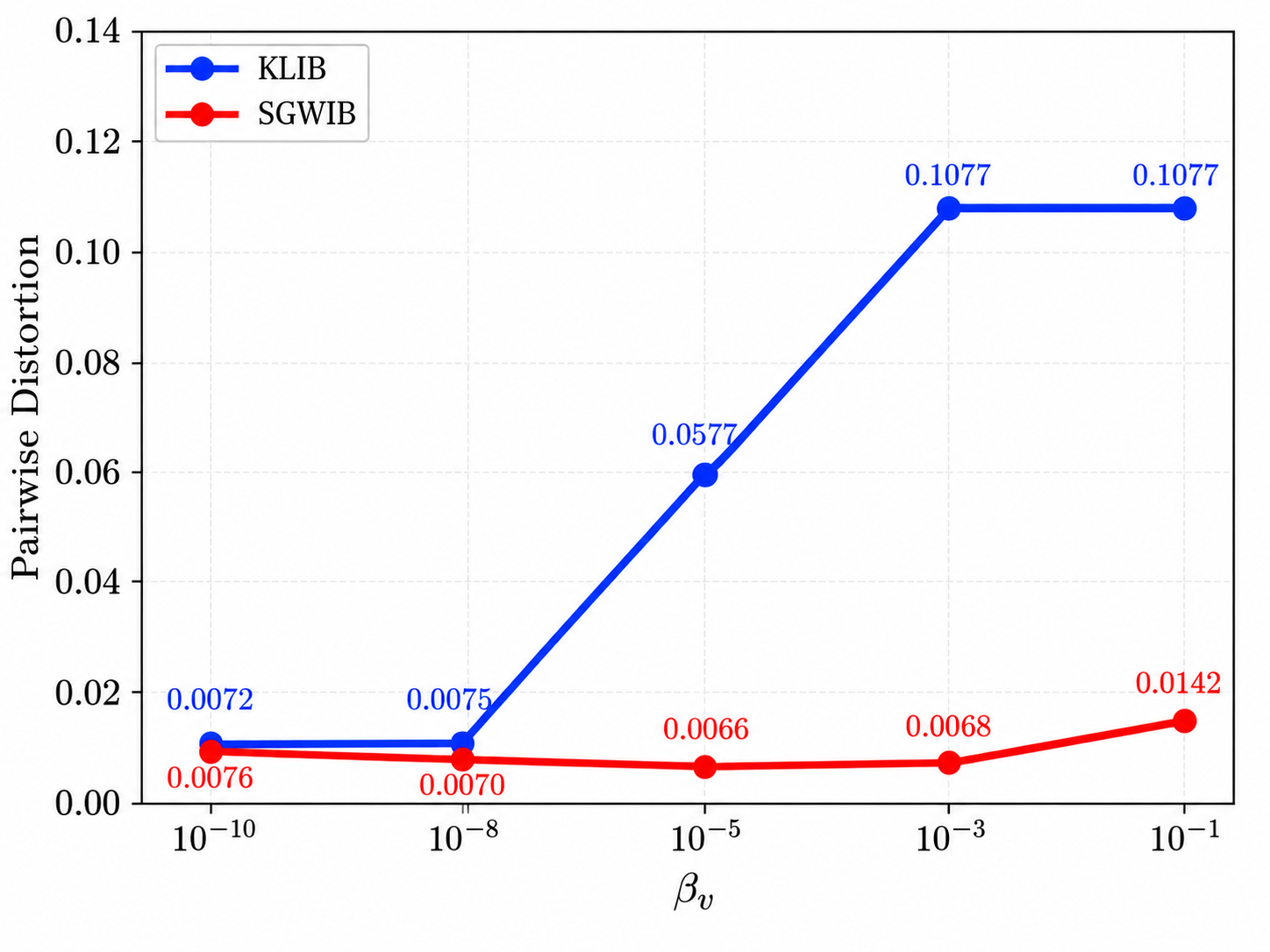}
    }
    \hfill
    \subfloat[
        Paired shift
        $\mathcal D_{\mathrm{paired}}^{v}$.
        \label{fig:paired_shift}
    ]{
        \includegraphics[width=0.31\textwidth]
        {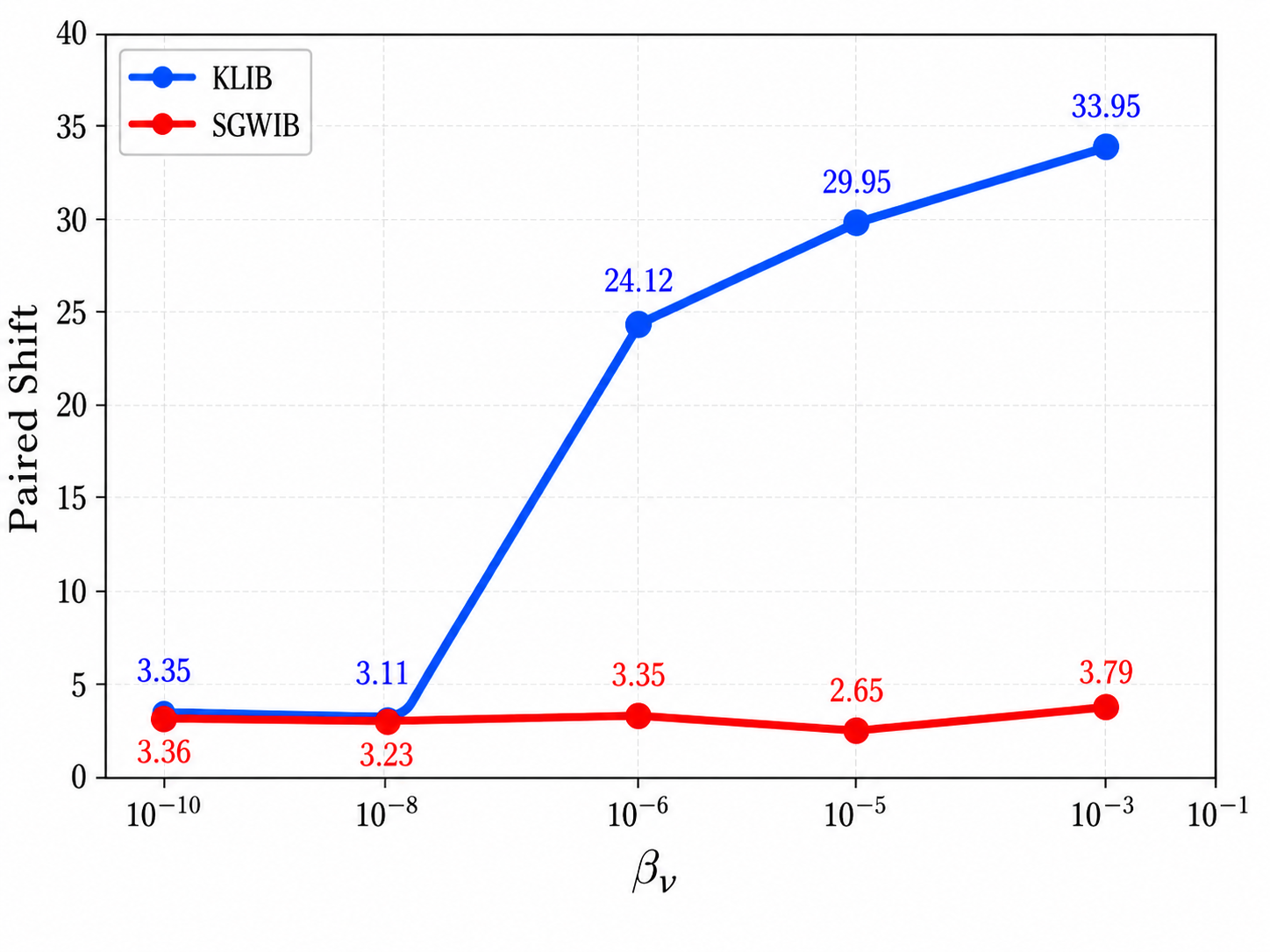}
    }
    \hfill
    \subfloat[
        Centroid shift
        $\mathcal D_{\mathrm{centroid}}^{v}$.
        \label{fig:centroid_shift}
    ]{
        \includegraphics[width=0.31\textwidth]
        {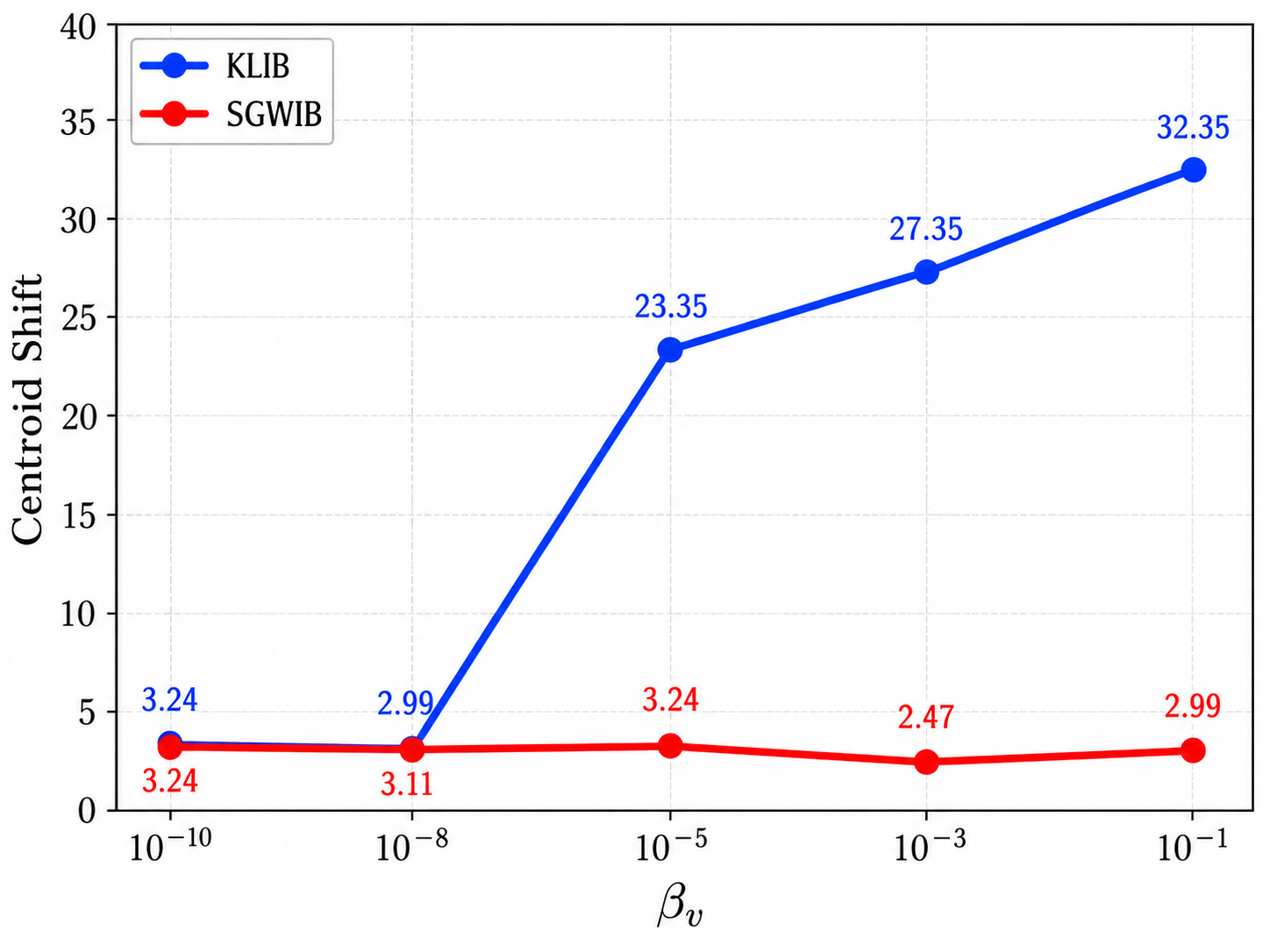}
    }
    \caption{Structural changes induced by KLIB and SGWIB under
    different bottleneck coefficients $\beta_v$ on the MrHiSum visual
    branch. Lower values indicate better structural preservation.}
    \label{fig:structural_preservation_metrics}
\end{figure*}

\begin{figure}[!t]
  \centering
  \includegraphics[width=0.95\columnwidth]{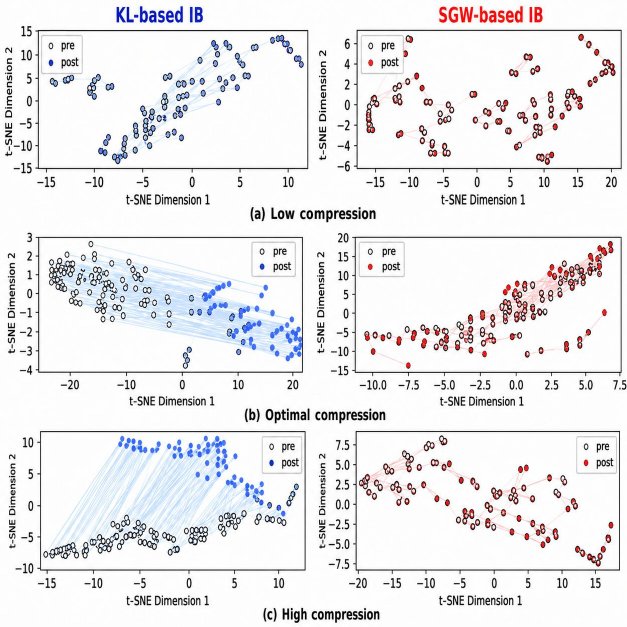}
  \caption{Joint t-SNE visualization of the input and bottleneck visual
  representations produced by KLIB and SGWIB on MrHiSum under (a) low
  ($\beta_v=10^{-10}$), (b) validation-selected ($\beta_v=10^{-6}$),
  and (c) high ($\beta_v=10^{-1}$) compression strengths. Open and
  filled circles denote $\widetilde{\mathbf x}_{t}^{v,\mathrm{std}}$
  and $\mathbf z_{t}^{v,\mathrm{std}}$, respectively, and each line
  connects the representations of the same valid temporal segment
  before and after compression.}
  \label{fig:tsne_structural_preservation}
\end{figure}

Figure~\ref{fig:structural_preservation_metrics} shows that the two
regularizers are indistinguishable under weak compression: at
$\beta_v\leq10^{-8}$, all three metrics stay near their lower bounds
for both methods. They separate sharply at $\beta_v=10^{-6}$, the
validation-selected setting for the MrHiSum visual branch, where KLIB
jumps from $0.0076$ to $0.0577$ in pairwise distortion, from $3.11$ to
$24.12$ in paired shift, and from $2.99$ to $23.35$ in centroid shift,
and keeps growing to $0.1077$, $33.95$, and $32.35$ at
$\beta_v=10^{-1}$. SGWIB instead remains within $0.0066$ to $0.0142$,
$2.65$ to $3.79$, and $2.47$ to $2.99$ across the same range.

Figure~\ref{fig:tsne_structural_preservation} illustrates the same
contrast. Under weak compression, both methods produce short and
unordered displacements between corresponding segments. Under the
validation-selected and high settings, KLIB transports the entire
segment set along a common direction, producing long parallel
trajectories, whereas the SGWIB embeddings keep each bottleneck vector
adjacent to its source counterpart. SGMG therefore preserves temporal
relational structure across the full range of $\beta_v$, while
KL-based prior matching does so only when the regularizer is too weak
to compress.
\subsubsection{Sensitivity to the Bottleneck Coefficient}
\label{subsubsec:beta_sensitivity}

We examine the modality-specific bottleneck coefficient $\beta_m$,
which controls the contribution of
$\mathcal L_{\mathrm{SGMG}}^{m}$ in
Eq.~\eqref{eq:total_training_loss}. On MrHiSum,
\[
\mathcal H_{\beta}
=
\left\{
10^{-10},
10^{-9},
10^{-8},
10^{-7},
10^{-6},
10^{-5}
\right\}.
\]
Smaller $\beta_m$ emphasizes highlight-score prediction, whereas
larger values impose stronger structure-aware compression.

\begin{figure*}[!t]
    \centering

    \subfloat[
        Kendall's $\tau$.
        \label{fig:beta_tau}
    ]{
        \includegraphics[width=0.30\textwidth]
        {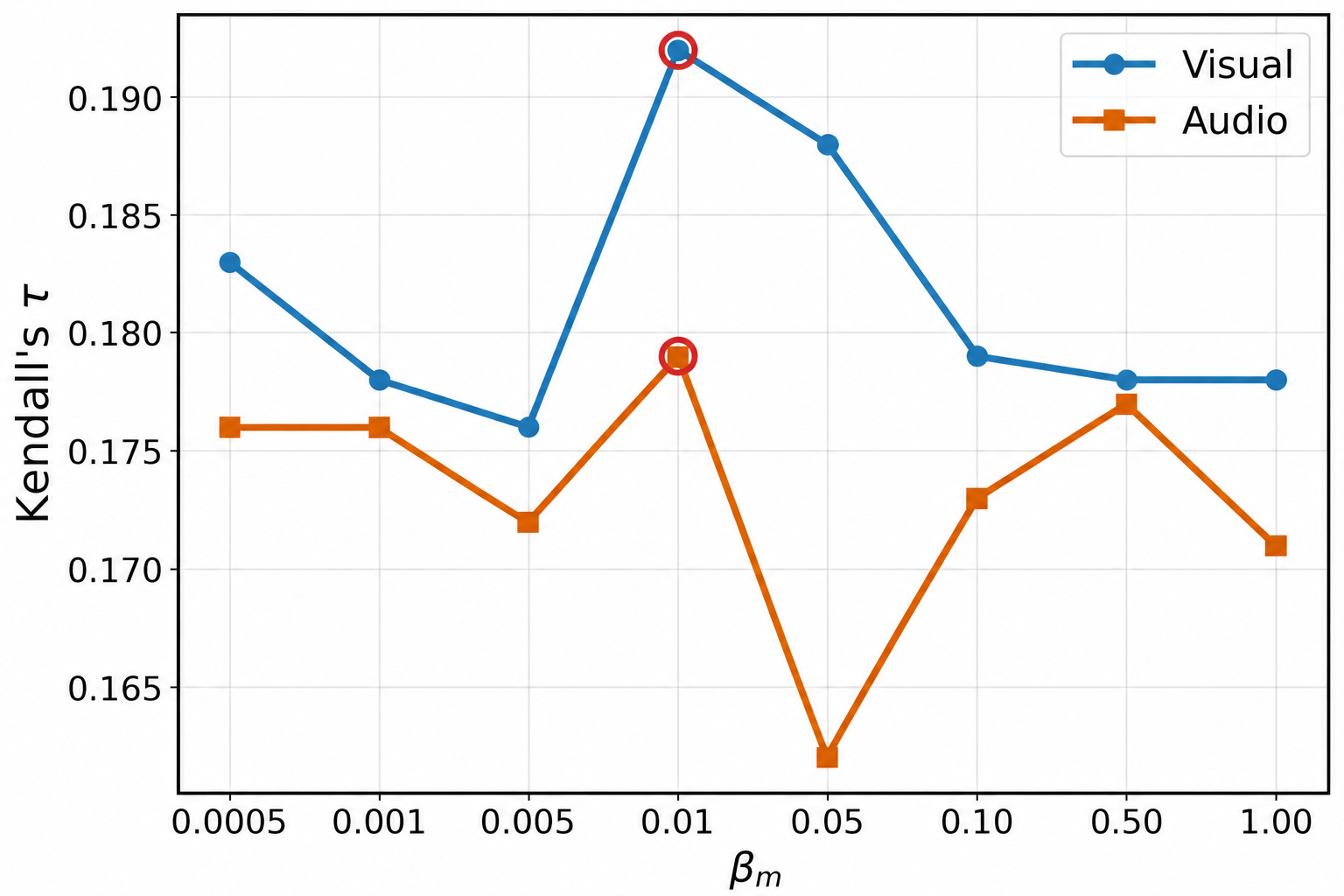}
    }
    \hfill
    \subfloat[
        Spearman's $\rho$.
        \label{fig:beta_rho}
    ]{
        \includegraphics[width=0.30\textwidth]
        {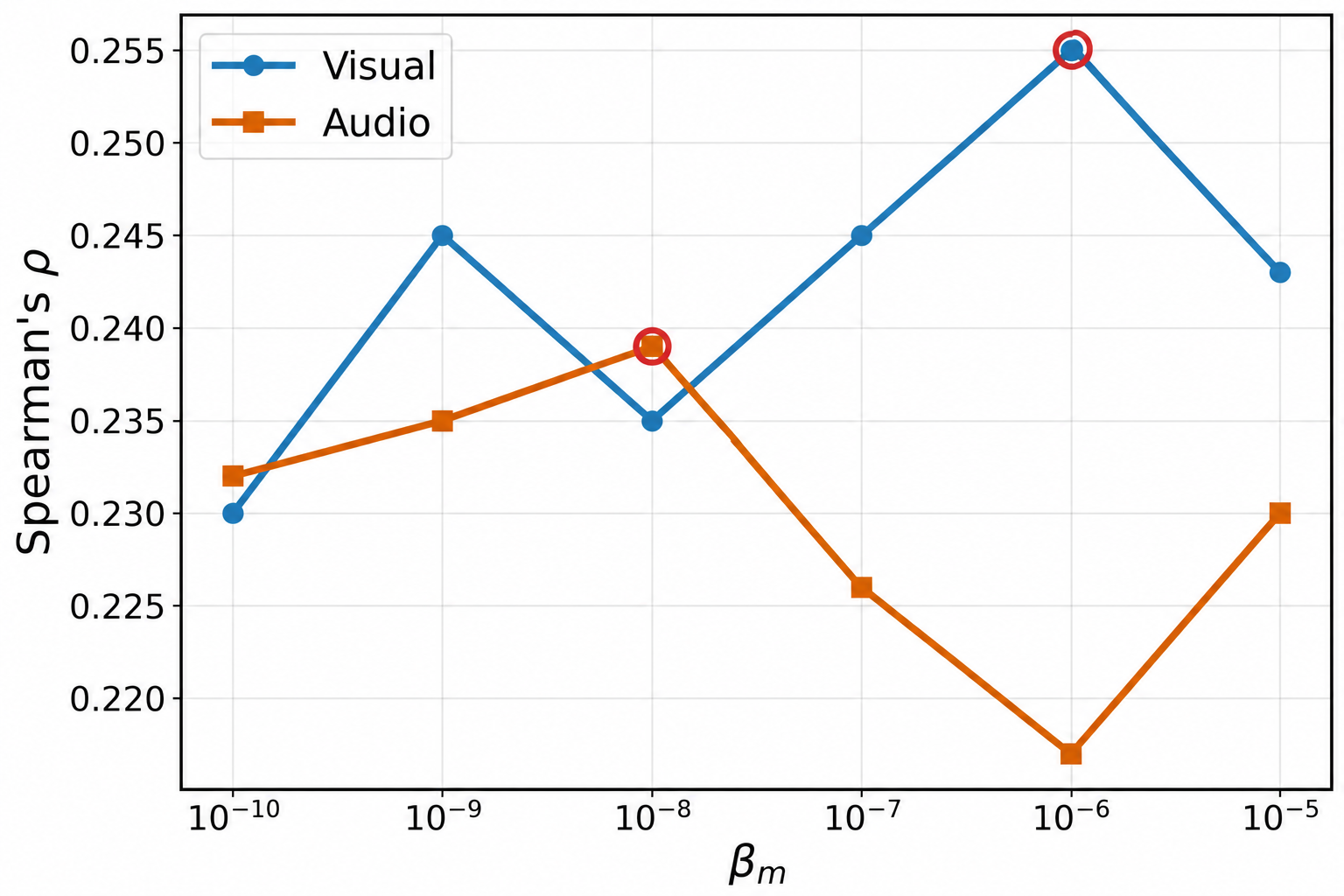}
    }
    \hfill
    \subfloat[
        $\mathrm{mAP}@50$.
        \label{fig:beta_map50}
    ]{
        \includegraphics[width=0.30\textwidth]
        {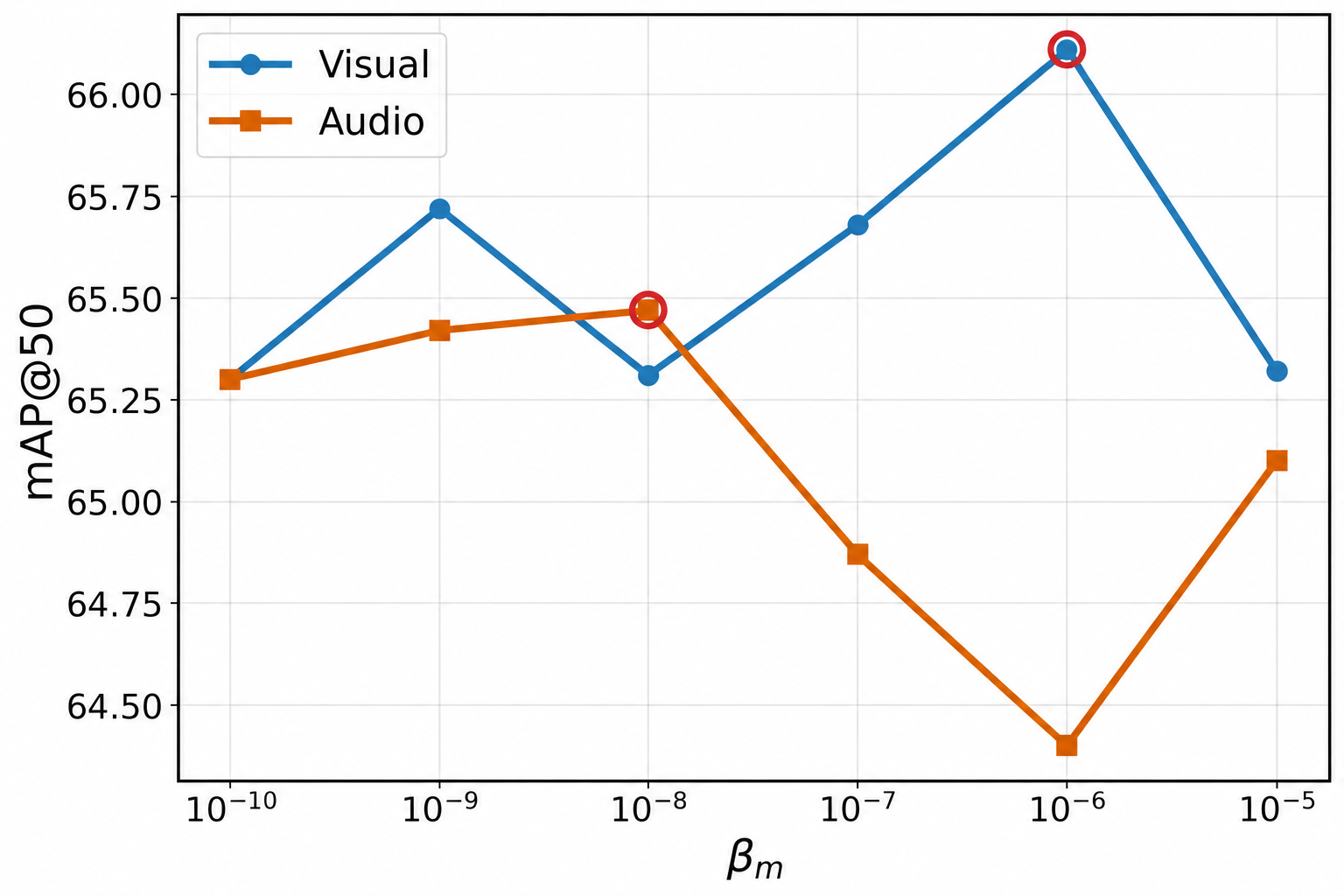}
    }

    \vspace{0.5em}

    \subfloat[
        $\mathrm{mAP}@30$.
        \label{fig:beta_map30}
    ]{
        \includegraphics[width=0.30\textwidth]
        {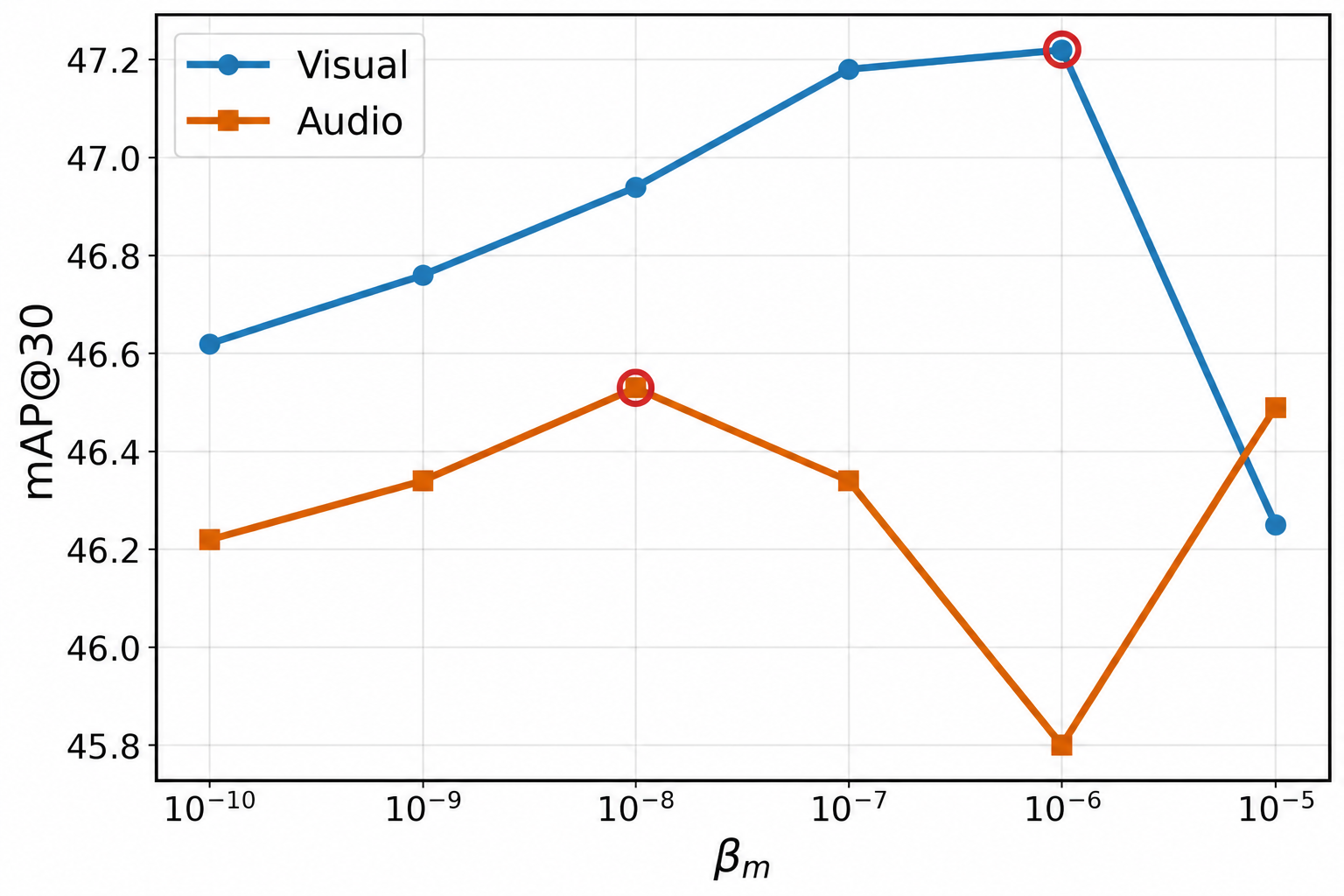}
    }
    \hspace{0.04\textwidth}
    \subfloat[
        $\mathrm{mAP}@15$.
        \label{fig:beta_map15}
    ]{
        \includegraphics[width=0.30\textwidth]
        {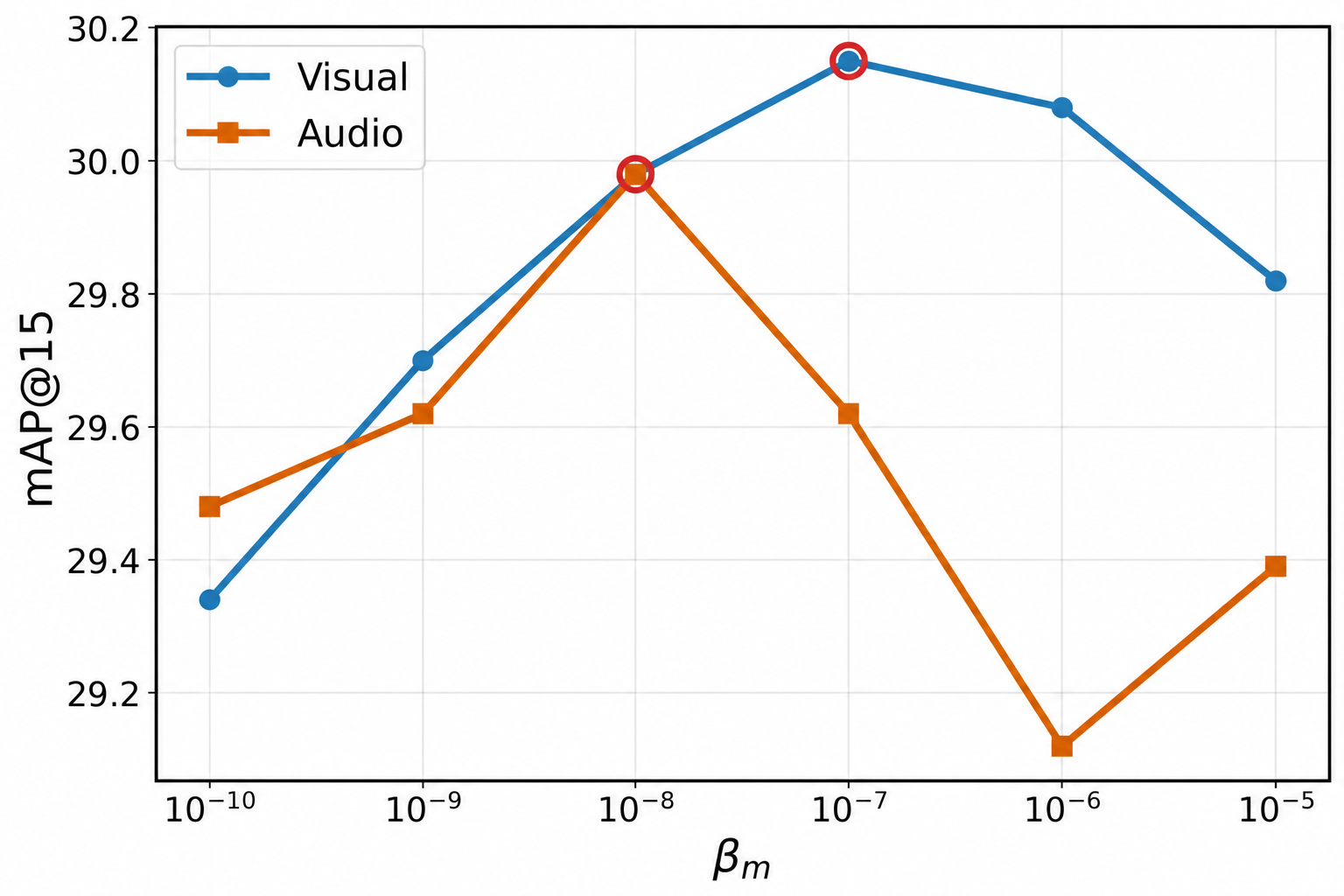}
    }

    \caption{Sensitivity of the independently trained visual and audio
    SGWIB models to the bottleneck coefficient $\beta_m$ on MrHiSum.
    Each panel compares the two modalities under the same evaluation
    metric. Red circles indicate the best value of each modality for
    the corresponding metric.}
    \label{fig:beta_sensitivity}
\end{figure*}

As shown in Fig.~\ref{fig:beta_sensitivity}, performance varies
non-monotonically with $\beta_m$, and the two modalities peak at
different coefficients. The visual branch is strongest around
$\beta_v=10^{-6}$, which is the best setting for four of the five
metrics, while the audio branch consistently peaks at
$\beta_a=10^{-8}$ on all five. The two curves move in opposite
directions in between: at $\beta=10^{-6}$ the visual branch reaches its
maximum whereas the audio branch falls to its minimum on every metric.
Excessively weak regularization provides limited structural constraint,
while overly strong compression suppresses highlight-relevant
information, and the offset between the two optima supports the use of
modality-specific bottleneck coefficients.

\subsubsection{Sensitivity to the HAR-CDM Loss Weight}
\label{subsubsec:gamma_sensitivity}

We further examine the HAR-CDM coefficient $\gamma_m$ while fixing the
bottleneck coefficients to their validation-selected values,
$\beta_v^\star=10^{-6}$ and $\beta_a^\star=10^{-8}$.

\begin{figure*}[!t]
    \centering

    \subfloat[
        Kendall's $\tau$.
        \label{fig:gamma_tau}
    ]{
        \includegraphics[width=0.30\textwidth]
        {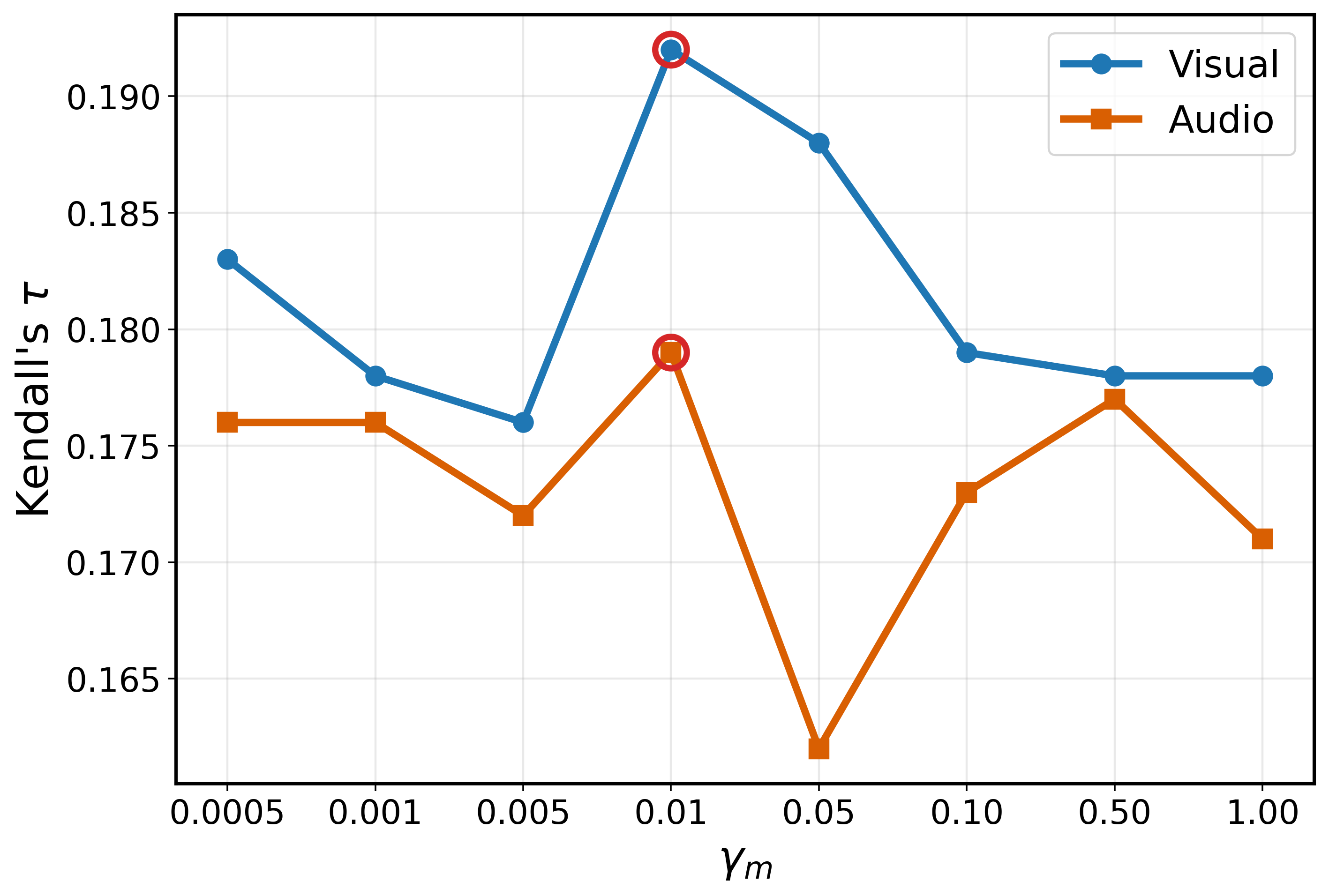}
    }
    \hfill
    \subfloat[
        Spearman's $\rho$.
        \label{fig:gamma_rho}
    ]{
        \includegraphics[width=0.28\textwidth]
        {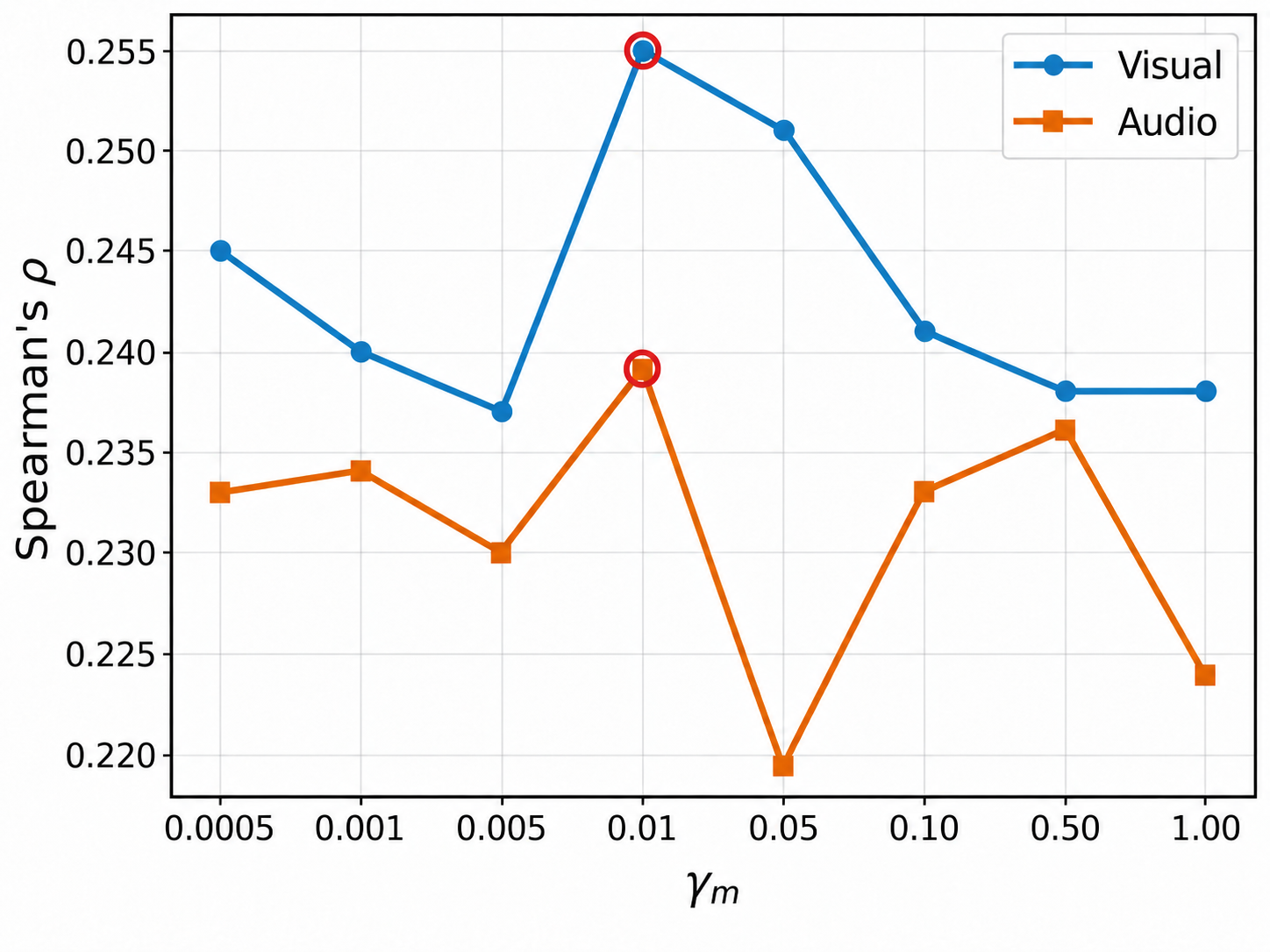}
    }
    \hfill
    \subfloat[
        $\mathrm{mAP}@50$.
        \label{fig:gamma_map50}
    ]{
        \includegraphics[width=0.28\textwidth]
        {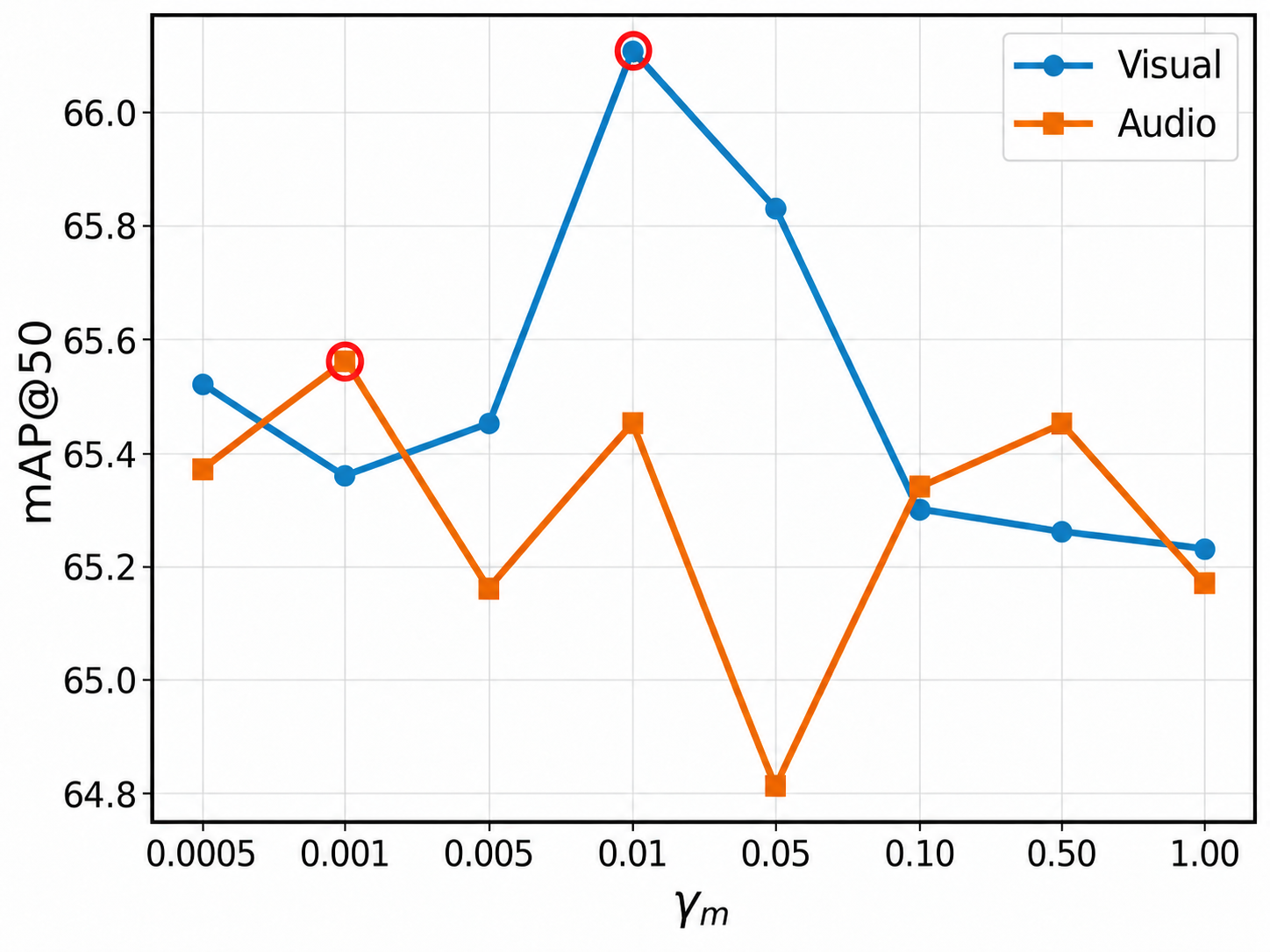}
    }

    \vspace{0.5em}

    \subfloat[
        $\mathrm{mAP}@30$.
        \label{fig:gamma_map30}
    ]{
        \includegraphics[width=0.28\textwidth]
        {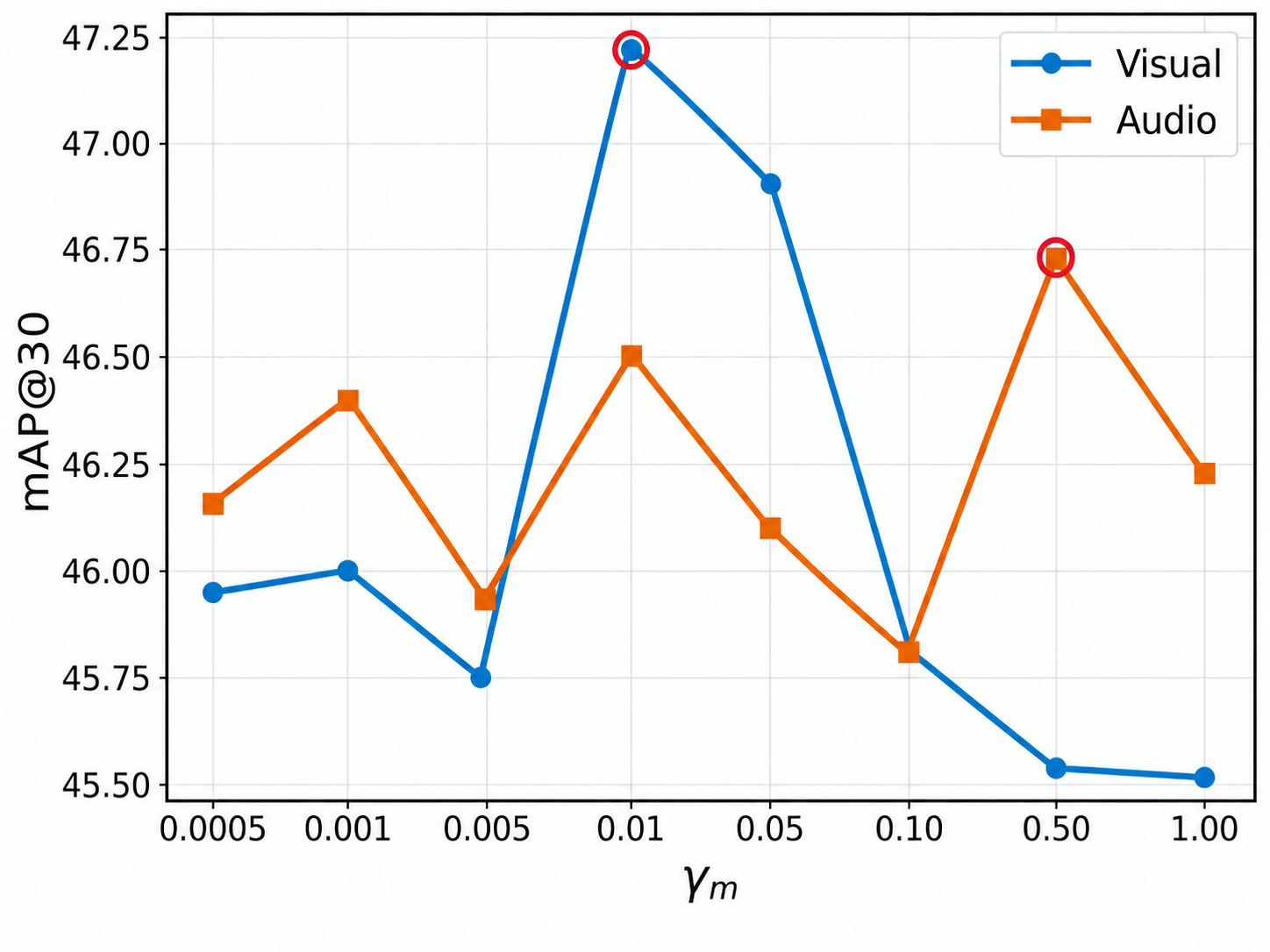}
    }
    \hspace{0.04\textwidth}
    \subfloat[
        $\mathrm{mAP}@15$.
        \label{fig:gamma_map15}
    ]{
        \includegraphics[width=0.28\textwidth]
        {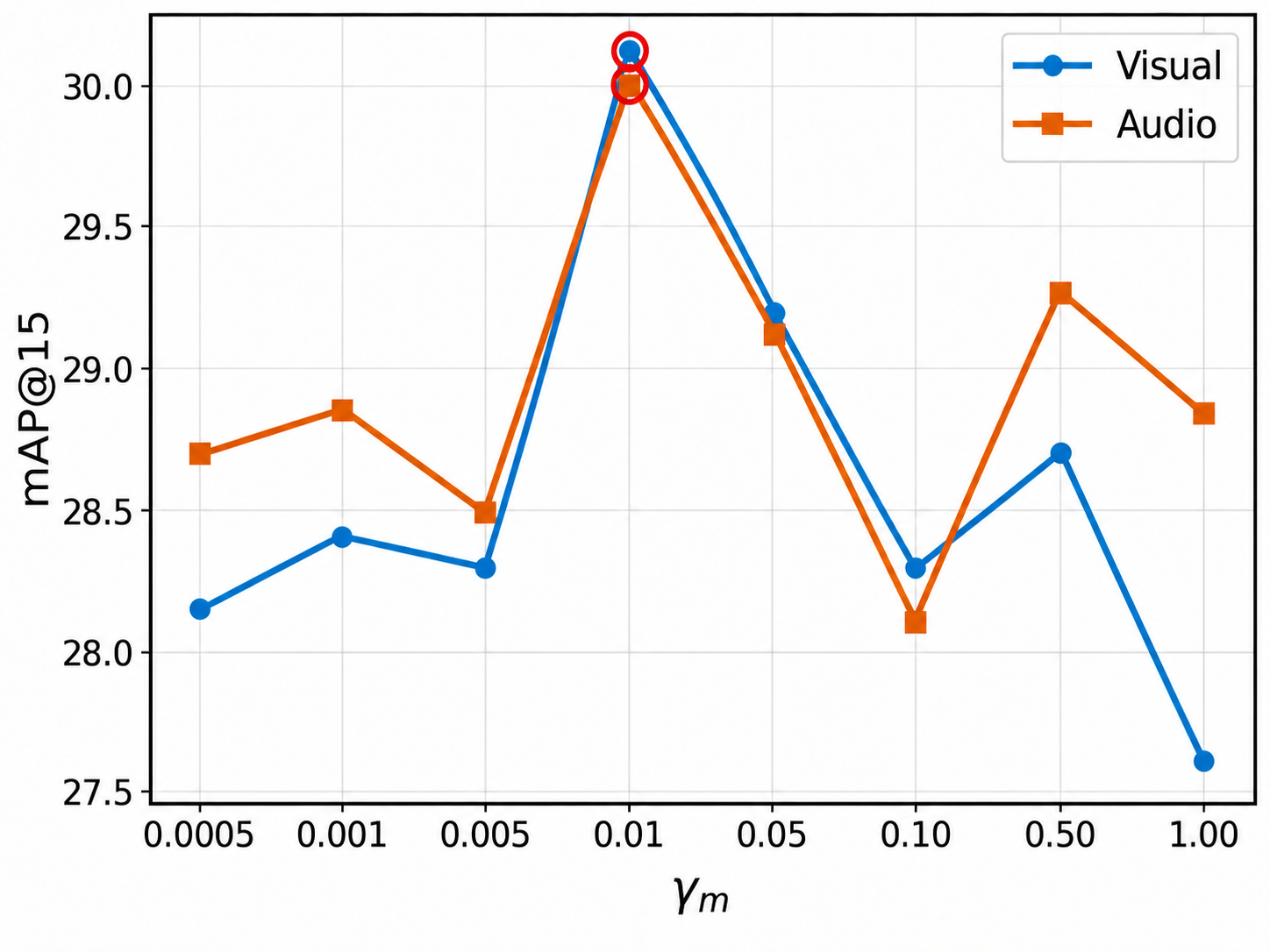}
    }

    \caption{Sensitivity to the HAR-CDM loss weight $\gamma_m$ on
    MrHiSum. Visual and audio denote independently trained single-modal
    models. The bottleneck coefficients are fixed to
    $\beta_v^\star=10^{-6}$ and $\beta_a^\star=10^{-8}$,
    respectively. Red circles indicate the best value of each modality
    for the corresponding evaluation metric.}
    \label{fig:gamma_sensitivity}
\end{figure*}
As shown in Fig.~\ref{fig:gamma_sensitivity}, the visual branch exhibits
a clear non-monotonic response, with all five metrics peaking at
$\gamma_v=0.01$. The audio branch is more metric dependent: Kendall's
$\tau$, Spearman's $\rho$, and $\mathrm{mAP}@15$ also favor
$\gamma_a=0.01$, whereas $\mathrm{mAP}@50$ peaks at $0.001$ and
$\mathrm{mAP}@30$ at $0.5$. Both modalities nonetheless degrade sharply
at $\gamma_m=0.05$, indicating that the sensitivity is driven by the
strength of the disentanglement constraint rather than by metric-specific
noise. Overall, a moderate HAR-CDM weight is preferable for both
modalities, with the audio branch tolerating a wider range, which is
consistent with HAR-CDM and SGWIB playing complementary roles in
task/context disentanglement and structural compression.

\section{Conclusion}
\label{sec:conclusion}

This paper proposed SGWIB, a structure-aware information bottleneck
framework for video highlight detection. By introducing the Sliced
Gromov--Monge Gap (SGMG), SGWIB preserves temporal relational
structure during compression, while HAR-CDM further separates
task-oriented and context-oriented information. Experiments on
MrHiSum and MoSu demonstrate consistent improvements in temporal
ranking and highlight retrieval, and ablation studies confirm the
complementary effects of the two components. Compared with
conventional KLIB, SGWIB achieves better task performance and
substantially stronger structural preservation with only modest
additional computational cost.Future work will focus on finer-grained highlight selection, adaptive
structural bottlenecks, and extensions to multimodal video
representation learning.
\section*{Acknowledgment}

The research is partially supported by the Ministry of Science and
Technology grant MOST 109-2221-E-011-127-MY3, and National
Science and Technology Council grants NSTC 112-2221-E-011-111, NSTC 112-2634-F-011-002-MBK,  NSTC 113-2221-E-011-119, NSTC 113-2634-F-011-002-MBK, NSTC 113-2221-E-011-119, NSTC 113-2634-F-011-002-MBK, NSTC 114-2221-E-011-058-MY3, and NSTC 114-2634-F-011-002-MBK, Fujian Natural Science Foundation (2025J011063,2025R0161), and Wuyi University Talent Introduction Research Startup Project (YJ202512).

Generative AI tools were used solely for language editing and
grammatical refinement. All technical content, analyses, and
conclusions were developed and verified by the authors.

\bibliographystyle{IEEEtran}
\bibliography{references}

\end{document}